%% file: main.tex
\documentclass{article} 
\usepackage{iclr2023_conference,times}

\input{math_commands.tex}

\usepackage{hyperref}
\usepackage{url}
\usepackage{enumitem}
\usepackage{multirow}
\usepackage{amsmath}
\usepackage{mathtools}
\usepackage{graphicx}
\usepackage{booktabs}
\usepackage{subcaption}
\usepackage[ruled, noend]{algorithm2e}
\usepackage{animate}
\usepackage{tikz}
\usepackage{comment}
\usepackage{amssymb} %
\usepackage{color}
\usepackage{wrapfig}
\usepackage{tabularx}
\usepackage{subcaption}

\hypersetup{
    colorlinks=true,
    citecolor=teal,
    linkcolor=red,
    urlcolor=teal
}

\title{\ours: A 3D Generative Model for Portrait Video Generation}

\author{Zhongcong Xu$^{1*}$,\quad Jianfeng Zhang$^{2}$\thanks{Equal contribution, work done during an internship at ByteDance} ,\quad Jun Hao Liew$^{2}$,\quad Wenqing Zhang$^{2}$, \\
{\bf Song Bai}$^{2}$,\quad {\bf Jiashi Feng}$^{2}$,\quad {\bf Mike Zheng Shou}$^{1}$\thanks{Corresponding author}\\
$^1$ Show Lab, National University of Singapore\hspace{1.0em}$^2$ ByteDance\\
\texttt{\{zhongcongxu,zhangjianfeng\}@u.nus.edu}\\
\texttt{\{junhao.liew,wenqingzhang,jshfeng\}@bytedance.com}\\
\texttt{\{songbai.site, mike.zheng.shou\}@gmail.com}}

\newcommand{\ie}{{\it i.e.}}
\newcommand{\eg}{{\it e.g.}}
\newcommand{\ours}{PV3D}

\iclrfinalcopy 
\begin{document}

\maketitle

\vspace{-5mm}
\begin{abstract}

Recent advances in generative adversarial networks (GANs) have demonstrated the capabilities of generating stunning photo-realistic portrait images. While some prior works have applied such image GANs to unconditional 2D portrait video generation and static 3D portrait synthesis, there are few works successfully extending GANs for generating 3D-aware portrait videos.
In this work, we propose \ours{}, the first generative framework that can synthesize multi-view consistent portrait videos. 
Specifically, our method extends the recent static 3D-aware image GAN to the video domain by generalizing the 3D implicit neural representation to model the spatio-temporal space. 
To introduce motion dynamics into the generation process, we develop a motion generator by stacking multiple motion layers to synthesize motion features via modulated convolution. 
To alleviate motion ambiguities caused by camera/human motions, we propose a simple yet effective camera condition strategy for \ours{}, enabling both temporal and multi-view consistent video generation. 
Moreover, \ours{} introduces two discriminators for regularizing the spatial and temporal domains to ensure the plausibility of the generated portrait videos. 
These elaborated designs enable \ours{} to generate 3D-aware motion-plausible portrait videos with high-quality appearance and geometry, significantly outperforming prior works.
As a result, \ours{} is able to support downstream applications such as static portrait animation and view-consistent motion editing. Code and models are available at  \href{https://showlab.github.io/pv3d}{https://showlab.github.io/pv3d}.

\end{abstract}

\input{Intro}

\input{Related_work}

\input{Method}

\input{Experiment}

\input{Conclusion}

\input{Acknowledgements}

\input{Ethnics_statement}

\input{Reproducibility_statement}

\bibliography{reference}
\bibliographystyle{iclr2023_conference}

\newpage
\appendix
\input{Appendix}

\end{document}

%% file: math_commands.tex
\usepackage{amsmath,amsfonts,bm}

\def\eqref#1{equation~\ref{#1}}

\def\1{\bm{1}}

\DeclareMathAlphabet{\mathsfit}{\encodingdefault}{\sfdefault}{m}{sl}
\SetMathAlphabet{\mathsfit}{bold}{\encodingdefault}{\sfdefault}{bx}{n}



%% file: Intro.tex
\section{Introduction}


Recent progress in generative adversarial networks (GANs) has led human portrait generation to unprecedented success~\citep{karras2020analyzing,karras2021alias,skorokhodov2021stylegan} and has spawned a lot of industrial applications~\citep{tov2021designing,richardson2021encoding}. Generating portrait videos has emerged as the next challenge for deep generative models 
with wider applications like video manipulation~\citep{abdal2022video2stylegan} and animation~\citep{siarohin2019first}. A long line of work has been proposed to either learn a direct mapping from latent code to portrait video~\citep{vondrick2016generating,saito2017temporal} or decompose portrait video generation into two stages, \ie, content synthesis and motion generation~\citep{tian2021good,tulyakov2018mocogan,skorokhodov2021stylegan}.

Despite offering plausible results, such methods only produce 2D videos without considering the underlying 3D geometry, 
which is the most desirable attribute with broad applications such as portrait reenactment~\citep{doukas2021headgan}, talking face animation~\citep{siarohin2019first}, and VR/AR~\citep{cao2022authentic}. Current methods typically create 3D portrait videos through {classical graphics techniques}~\citep{wang2021learning, ma2021pixel,grassal2022neural}, which require multi-camera systems, well-controlled studios, and heavy artist works. 
{In this work, we aim to alleviate the effort of creating high-quality 3D-aware portrait videos {by learning from 2D monocular videos only, \textbf{without} the need of any 3D or multi-view annotations}.}
\begin{figure}[ht]
    \centering
    \includegraphics[width=0.6\linewidth]{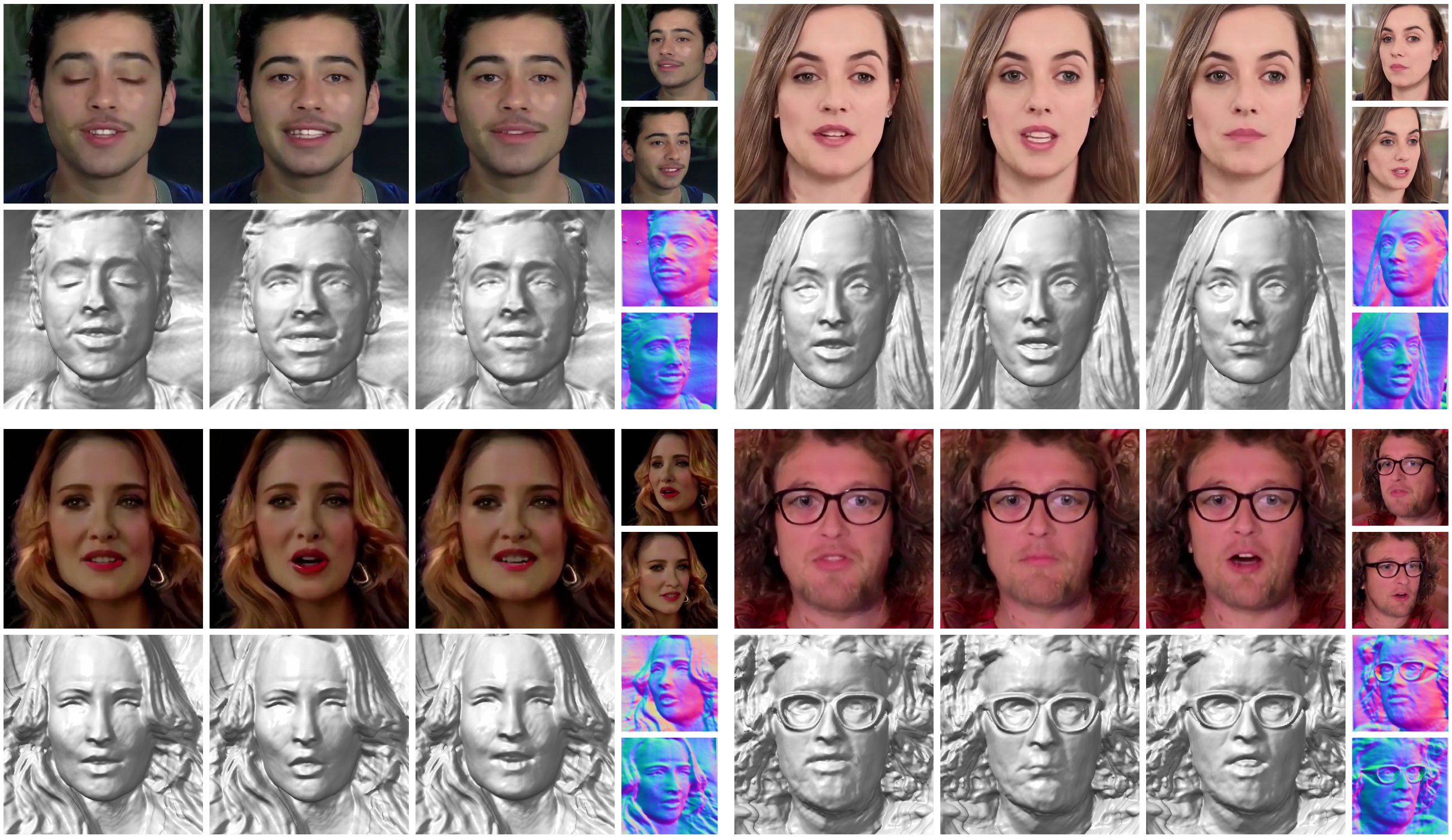}
    \caption{Our \ours{} can generate photo-realistic portrait videos with diverse motions and dynamic 3D geometry. We render surfaces extracted by marching cubes. The video frames and shape (normal map) can be rendered from arbitrary viewpoints. Please see our \href{https://showlab.github.io/pv3d}{project page} for video results. 
}
\vspace{-0.5em}
\label{fig:teaser}
\end{figure}

Recent 3D-aware portrait generative methods have witnessed rapid advances~\citep{schwarz2020graf,gu2021stylenerf,chan2021pi,niemeyer2021giraffe,or2021stylesdf,chan2022efficient}. Through integrating implicit neural representations (INRs)~\citep{sitzmann2020implicit, mildenhall2020nerf} into GANs~\citep{karras2019style,karras2020analyzing}, they can produce photo-realistic and multi-view consistent results.
{However, such methods are limited to static portrait generation and can hardly be extended to portrait video generation due to several challenges: 
1) it remains unclear how to effectively model 3D dynamic human portrait in a generative framework; 
2) learning dynamic 3D geometry without 3D supervision is highly under-constrained; 3) entanglement between camera movements and human motions/expressions introduces ambiguities to the training process.} {To this end, we propose a {\bf 3D} {\bf P}ortrait {\bf V}ideo generation model (\ours{}), the first method that can generate high-quality 3D portrait videos with diverse motions while learning purely from monocular 2D videos.} \ours{} enables 3D portrait video modeling by extending 3D tri-plane representation~\citep{chan2022efficient} to the spatio-temporal domain. \textcolor{black}{In this paper, we comprehensively analyze various design choices and arrive at a set of novel designs, including decomposing latent codes into appearance and motion components, temporal tri-plane based motion generator, proper camera pose sequence conditioning, and camera-conditioned video discriminators, which can significantly improve the video fidelity and geometry quality for 3D portrait video generation.}

As shown in Figure~\ref{fig:teaser}, despite being trained from only monocular 2D videos, \ours{} can generate a large variety of photo-realistic portrait videos under arbitrary viewpoints with diverse motions and high-quality 3D geometry.
Comprehensive experiments on various datasets including VoxCeleb~\citep{nagrani2017voxceleb}, CelebV-HQ~\citep{zhu2022celebvhq} and TalkingHead-1KH~\citep{wang2021facevid2vid} well demonstrate the superiority of \ours{} over previous state-of-the-art methods, both qualitatively and quantitatively. 
{Notably, it achieves 29.1 FVD on VoxCeleb, improving upon a concurrent work 3DVidGen~\citep{bahmani20223d} by 55.6\%. \ours{} can also generate high-quality 3D geometry, achieving the best multi-view identity similarity and warping error across all datasets.}

Our contributions are three-fold. 1) To our best knowledge,  \ours{} is the first method that is capable to generate a large variety of 3D-aware portrait videos with high-quality appearance, motions, and geometry. 2) \textcolor{black}{We propose a novel temporal tri-plane based video generation framework that can synthesize 3D-aware portrait videos by learning from 2D videos only.} 3) We demonstrate state-of-the-art 3D-aware portrait video generation on three datasets. Moreover, our \ours{} supports several downstream applications, \ie, static image animation, monocular video reconstruction, and multi-view consistent motion editing.

%% file: Related_work.tex
\section{Related Work}

{\bf 2D video generation.}
Early video generation works~\citep{vondrick2016generating,saito2017temporal} propose to learn a video generator to transform random vectors directly to video clips. While recent video generation works adopt a similar paradigm to design the video generator, \ie, disentangle the video content and motion (trajectory), then control them by different random noises. For the video content, most of the works build their frameworks on top of generative adversarial networks (GAN) designed for image domain, such as StyleGAN~\citep{karras2019style} and INR-GAN~\citep{skorokhodov2021adversarial}. Based on image GANs, video GAN works further extend the generation process to temporal domain using various motion generation approaches. MoCoGAN~\citep{tulyakov2018mocogan} and its following work MoCoGAN-HD~\citep{tian2021good} generate the motion code sequence autoregressively, which is implemented as random process. StyleGAN-V~\citep{skorokhodov2021stylegan} also generates a motion code sequence for each frame, while the motion codes are sampled separately and thus its generator can synthesize video frames independently. In contrast, DiGAN~\citep{yu2022generating} only sample one motion code for the entire video and the motion for each frame is generated by the INR network using time instant. This compact video code design is in line with the property of temporal consistency, \eg, a talking person only move the lips or twinkles, while the face shape does not change rapidly~\citep{tewari2019fml}. 

{\bf 3D-aware generation.}
Image GANs~\citep{karras2020analyzing, karras2021alias} have demonstrated impressive capability in synthesizing high-resolution photo-realistic images. By incorporating implicit neural representations~\citep{mildenhall2020nerf, sitzmann2020implicit} or differentiable neural rendering~\citep{kato2018neural} into GANs, recent works~\citep{schwarz2020graf,niemeyer2021giraffe, chan2021pi, shi2021lifting} can produce multi-view consistent images.
However, most of them are limited at image quality and resolution due to the heavy computation cost of traditional 3D representations. \textcolor{black}{Follow-up works are proposed to address this issue by either developing an efficient 3D representation~\citep{chan2022efficient,deng2021gram,schwarz2022voxgraf,zhao2022generative}} or dividing image generation into two stages~\citep{gu2021stylenerf,or2021stylesdf, zhang2022multi}, \ie, generating low-resolution images by volume rendering~\citep{max1995optical} and then refine them using super-resolution approaches. However, these methods are limited to static object generation. \textcolor{black}{Recently, CoRF~\citep{zhuang2022controllable} proposes to condition the static 3D GAN on estimated motion features. In addition, a concurrent work 3DVidGen~\citep{bahmani20223d} has been proposed to extend unconditional 3D-aware generation into video domain.} Although 3DVidGen can generate plausible results with changing viewpoints, its video fidelity and multi-view consistency are unsatisfactory. Moreover, 3DVidGen does not present or evaluate 3D geometry quality, making its 3D geometry generation ability unclear, which hinders its applicability in the scenarios requiring good 3D geometry~\citep{yuan2022nerf}. {In contrast, our \ours{} aims to synthesize realistic videos with high-quality detailed geometry.}


%% file: Method.tex
\section{PV3D: 3D Portrait Video Generation}
\subsection{Overview}
{\bf Problem formulation.}
{Given a monocular 2D portrait video collection $\mathfrak{D}=\{\boldsymbol{v}_n\}_{n=1}^{N}$ consisting of $N$ video sequences, the goal of 3D-aware portrait video generation is to learn a generator $\mathcal{G}$ that synthesizes videos given joint conditions of random noise $\mathbf{z}$, camera viewpoint $c$ and timestep $t$ \textbf{without} relying on 3D geometry or multi-view supervision.}

\begin{figure}[h]
\vspace{-1em}
\centering
\includegraphics[width=0.75\textwidth]{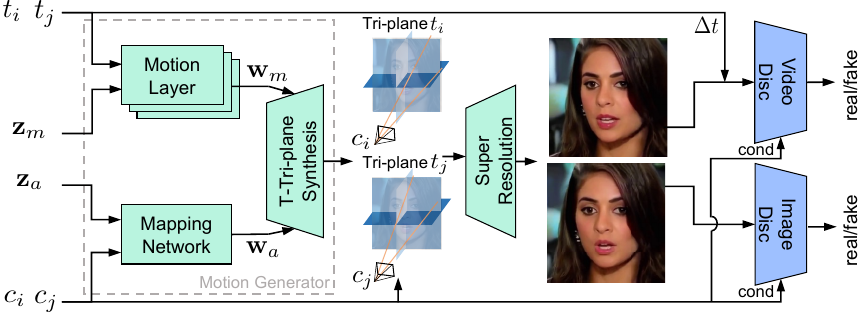}
\vspace{-3mm}
\caption{
\ours{} creates portrait video frames from appearance code $\mathbf{z}_a$, motion code $\mathbf{z}_m$, timesteps $\{t_i,t_j\}$, and camera poses $\{c_i,c_j\}$. Motion layers and mapping network encode inputs into intermediate motion code $\mathbf{w}_m$ and style code $\mathbf{w}_a$, respectively. 
To incorporate temporal dynamics, the temporal tri-plane synthesis network extends 3D tri-plane~\citep{chan2022efficient} to spatio-temporal domain.
Two camera-conditioned discriminators regularize the image quality and motion plausibility. 
}
\label{fig:arch}
\end{figure}

{\bf Framework.}
{The overview of our proposed framework is shown in Figure~\ref{fig:arch}. \ours{} formulates 3D-aware portrait video generation task as $\boldsymbol{v}=\mathcal{R}(\mathcal{G}(\mathbf{z}|c,t), c)$ where the generator $\mathcal{G}(\cdot)$ first generates 3D-aware spatio-temporal representation, followed by volume rendering and super-resolution (denoted as $\mathcal{R}(\cdot)$) to obtain the final video sequence. In this work, our generator $\mathcal{G}$ builds upon tri-plane representation from EG3D~\citep{chan2022efficient} and extends it to spatio-temporal representation for video synthesis, which we denote as temporal tri-plane.
Instead of jointly modeling appearance and motion dynamics within a single latent code $\mathbf{z}$, we factorize the 3D video generation into appearance and motion generation components. Specifically, our \ours{} takes two independent latent codes, \ie, appearance code $\mathbf{z}_a\sim \mathcal{N}(\mathbf{0},\boldsymbol{I})$ and motion code $\mathbf{z}_m\sim \mathcal{N}(\mathbf{0},\boldsymbol{I})$ as inputs.
We condition the generator on $\mathbf{z}_a$ to synthesize varying video appearance, \eg, genders, skin colors, hair styles, glasses, \textit{etc.}, and use $\mathbf{z}_m$ to model motion dynamics, \eg, a person opening his/her mouth.
}

During training, we {randomly} sample two timesteps $\{t_i$, $t_j\}$ and their corresponding camera poses $\{c_i$, $c_j\}$ for one video.
Following EG3D, we project the appearance code $\mathbf{z}_a$ and camera pose $c$ into intermediate appearance code $\mathbf{w}_a$ for content synthesis. 
As for the motion component, we develop motion layer to encode motion code $\mathbf{z}_m$ and timesteps $\{t_i, t_j\}$ into intermediate motion code $\mathbf{w}_m$. Our temporal tri-plane synthesis network generates tri-plane features based on $\mathbf{w}_a$ and $\mathbf{w}_m$. With the generated tri-plane at $\{t_i, t_j\}$, volume rendering~\citep{max1995optical} is applied to synthesize frames with camera pose $c_i$ and $c_j$, respectively. The rendered frames are then upsampled and refined by a super-resolution module. To ensure the fidelity and plausibility of the generated frame content and motion, we develop two discriminators $\mathcal{D}_{\text{img}}$ and $\mathcal{D}_{\text{vid}}$ to supervise the training of $\mathcal{G}$. Both $\mathcal{D}_{\text{img}}$ and $\mathcal{D}_{\text{vid}}$ are camera-conditioned, which can leverage 3D priors. 

\subsection{3D-aware Video Generator}
\textbf{Challenges.}
Existing 3D-aware image GANs can only model static scenes and it remains unclear how to model motion dynamics, such as topological changes of a scene~\citep{park2021hypernerf}. A straightforward approach is to directly combine the motion condition with latent code, camera pose, and timestep, and feed them to the encoder to generate 3D representations for video rendering. However, such a naive design cannot generate temporally consistent and motion-plausible portrait videos because the generation of video content and motion is highly entangled. Besides, learning 3D-aware portrait video appearance and geometry from monocular 2D videos only is highly under-constrained, making the model training difficult and generation quality poor.
Another challenge is that motions caused by camera pose changes (\eg, camera movements) and head pose changes (\eg, look up, turn left) are highly entangled. This introduces ambiguities to the training process, largely increasing model's learning difficulties.

Prior works typically incorporate motion features by either manipulating the latent code for pre-trained image generator~\citep{tian2021good} or simply conditioning 3D representation on latent code and timestep~\citep{bahmani20223d}. Nevertheless, such designs cannot guarantee temporal consistency and motion diversity.
{To address these challenges, we decouple the latent code into appearance and motion components and propose a motion generator to model temporal dynamics. Such design not only preserves the high-fidelity and multi-view consistency of each frame but also enables the synthesis of videos with temporal coherence and motion diversity. Secondly, we propose a camera conditioning strategy to alleviate the motion ambiguity issue and thus facilitates convergence.
}

\begin{wrapfigure}{r}{0.41\textwidth}
  \vspace{-1.38em}
  \begin{center}
    \includegraphics[width=0.4\textwidth]{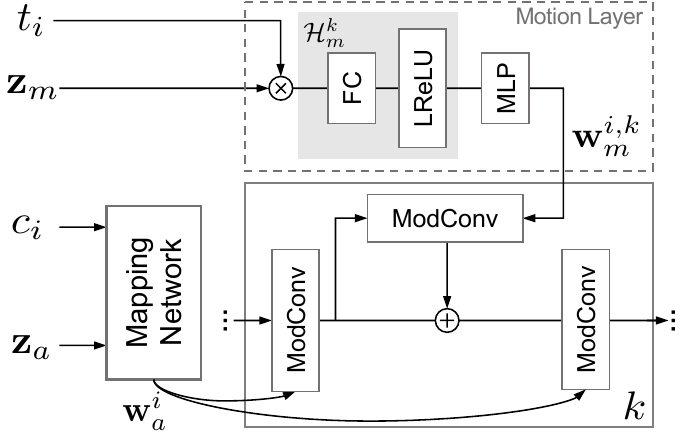}
    \vspace{-0.22em}
    \caption{Architecture of the $k$-th synthesis layer in motion generator. Motion layer encodes $\mathbf{z}_m$ and $t_i$ into intermediate motion code $\mathbf{w}_m^{i,k}$. Motion features are computed by modulating appearance features which are conditioned on $\mathbf{w}_a^{i}$.
    }
    \label{fig:motion_network}
  \end{center}
\end{wrapfigure}

\textbf{Synthesis of motion dynamics.}
To generate motion at each timestep in a video, we condition the generator on latent codes $\mathbf{z}_a$ and $\mathbf{z}_m$, timesteps $\{t_i$, $t_j\}$ and camera poses $\{c_i$, $c_j\}$. 
Without loss of generality, we only describe the generation process for timestep $t_i$ and camera pose $c_i$. As shown in Figure~\ref{fig:motion_network}, we introduce $K$ motion layers into the synthesis layers of motion generator. Each motion layer encodes motion code $\mathbf{z}_m$ and timestep $t_i$ into intermediate motion code $\mathbf{w}_m^{i,k}$. 
In particular, the motion code is first multiplied with timestep to encode temporal information, followed by a lightweight mapping head $\mathcal{H}_m$ with leaky ReLU activation. A multi-layer perceptron (MLP) then encodes them into $\mathbf{w}_m^{i,k}$. In other words, the $k$-th motion layer computes
\begin{equation}
    \mathbf{w}_m^{i, k} = \mathrm{MLP}_k(\mathcal{H}_m^{k}(\mathbf{z}_m*t_i))\text{,}
\end{equation}
where $k \in \{0, 1, ..., K\}$. $i \in \{0, 1, ..., N\}$ denotes the frame index while $N$ represents the total number of frames in one video. These intermediate motion codes $\mathbf{w}_m^{i,k}$ are then passed to the temporal tri-plane synthesis network to modulate the static appearance features via adaptive instance normalization (AdaIN)~\citep{karras2019style} to incorporate temporal dynamics.

In practice, we employ an equivalent operator, \ie, modulated convolution~\citep{karras2020analyzing}, to compute motion features. We then fuse it with the appearance features controlled by $\mathbf{w}_a^i$. The fused features are passed to the next synthesis layer iteratively to generate tri-planes. This process is formulated as
\begin{equation}
\textcolor{black}{\mathrm{f}_{k} = \mathcal{S}_k^{1}(\mathrm{f}^*+\mathrm{ModConv}(\mathrm{f}^*,\mathbf{w}_m^{i, k}), \mathbf{w}_a^i)\text{,  } \text{where}~\mathrm{f}^*=S_k^{0}(\mathrm{f}_{k-1}, \mathbf{w}_a^i)}
\text{.}
\end{equation}
Here, $\mathcal{S}_k^{0}$ and $\mathcal{S}_k^{1}$ denote the first and second synthesis block ($\mathrm{ModConv}$) in $k$-th synthesis layer, while $\mathrm{f}_{k}$ denotes the feature map synthesized by the $k$-th layer.
 
\textbf{Motion diversity and temporal consistency.}
Recent works~\citep{tov2021designing,shen2020interpreting,richardson2021encoding} investigate the semantic meaning of intermediate latent code space ($\mathcal{W}^{+}$) in pre-trained StyleGAN model, and discover that one can perform diverse manipulation of image content by using different style codes at different style-modulation layers. We made similar
observation for \ours{} (Appendix~\ref{appendix:wanalysis}). 
Specifically, mixing latent codes in shallow layers only brings coarse level changes on facial attributes, such as expressions, hair, \textit{etc}, which reduces the diversity. 
On the other hand, mixing style code in deeper layers leads to drastic changes in image content. As a result, identities will no longer be preserved, causing severe temporal inconsistency.

To preserve the identity in generated videos, we carefully select $K$ synthesis layers, such that $k\leq K$, for incorporating motion features. Choosing a suitable $K$ increases temporal consistency and improves our motion generator's capacity for modeling diverse motions. Besides, this synthesis layer selection also alleviates the overfitting of RGB video frames, which improves the quality of 3D geometry.

{\bf Alleviating motion ambiguities by camera sequence condition.} 
Learning 3D-aware portrait video generation from 2D videos faces another challenge, \ie, the entangled face motions and camera movements bring motion ambiguities. The concurrent work 3DVidGen~\citep{bahmani20223d} simply shares one camera pose for the entire video. However, this approach ignores motion ambiguity that harms video fidelity and geometry quality. Differently, we condition the generator on pre-estimated camera pose sequences (Appendix~\ref{appendix:process}). It has two advantages. 1) Suppose that we are only interested in the faces, a camera rotating around a static face is equivalent to rotating the face in front of a static camera. Generating head rotation directly is challenging due to the large topological changes of the scene. By conditioning our generator on $c_i$ at each time instant $t_i$, we can model the head rotation by rotating camera for observation instead of deforming the 3D scene, making our model easy to optimize. 2) Given the camera pose sequence, our generator can encode view-dependent features for each frame to leverage 3D priors. Such a simple design effectively improves the multi-view consistency and facilitates the learning of dynamics in 3D geometry as verified in our experiments.

\subsection{Conditional Discriminators}
Due to the lack of regularization in video generation procedure, the generated videos may present implausible contents or unreasonable motions. To ensure the plausibility of portrait video generation, we introduce a discriminator module to guide the generation process. In particular, the module consists of an image discriminator $\mathcal{D}_{\text{img}}$ for evaluating video appearance quality and a video discriminator $\mathcal{D}_{\text{vid}}$ for ensuring video motion plausibility. 

Our image discriminator $\mathcal{D}_{\text{img}}$ follows EG3D's discriminator architecture and uses camera poses as conditions to guide the generator to learn correct 3D priors and thus it can produce multi-view consistent portraits. We apply $\mathcal{D}_{\text{img}}$ on each generated frame $I_{i}$ (at timestep $t_i$) independently, which can be formulated as $p_{\text{img}} = \mathcal{D}_{\text{img}}(I_{i}, c_{i})$, where $p_{\text{img}}$ denotes the real/fake probability.

Our \ours{} generates two images $\{I_{{i}}, I_{{j}}\}$ jointly at two random timesteps $\{t_i, t_j\}$ for each video during training, we thus design a camera-conditioned dual-frame video discriminator $\mathcal{D}_{\text{vid}}$ to facilitate motion-plausible portrait video generation. Specifically, we first concatenate the generated two frames $\{I_{{i}}, I_{{j}}\}$ channel-wisely to obtain an image pair. To help encode temporal information, we further concatenate the timestep difference $\Delta t=t_j-t_i$ with the image pair. Our video discriminator learns to differentiate the real and generated image pairs based on motion features extracted from this hybrid input. To alleviate motion ambiguity and model view-dependent effects, we further condition $\mathcal{D}_{\text{vid}}$ on the corresponding camera poses $\{c_i, c_j\}$. Our video discriminator is formulated as $p_{\text{vid}} = \mathcal{D}_{\text{vid}}([I_{{i}},I_{{j}}, \Delta t], [c_i, c_j])$, where $p_{\text{vid}}$ indicates the probability of each image pair being sampled from real data distribution. Although $\mathcal{D}_{\text{vid}}$  only takes two frames as inputs, it can effectively learn ordinal information (Appendix~\ref{appendix:exp}) and help produce motion-plausible results as verified in our experiments (Section~\ref{exp:ab}). Moreover, such a simple design largely improves training efficiency and stability compared with previous methods that take {long sequences as conditions~\citep{tulyakov2018mocogan,tian2021good}}.

\subsection{Training and Inference}
{\bf Training.} 
To synthesize an image $I\in\mathbb{R}^{H\times W\times 3}$ based on tri-plane $\mathcal{T}$, we shoot rays $\boldsymbol{r}(s)=\boldsymbol{o}+s \boldsymbol{d}$ out from the camera origin $\boldsymbol{o}$ along direction $\boldsymbol{d}$ at each pixel~\citep{mildenhall2020nerf}. In practice, we sample query points $\boldsymbol{x}_{\boldsymbol{r}}$ along each ray and get the features for each point by interpolating them in $\mathcal{T}$. The features are passed to the decoder to predict color $\mathbf{c}$ and density $\sigma$ such that $[\sigma(\mathbf{r}(s), \mathbf{c}(\mathbf{r}(s))]=\mathrm{Decoder}(\mathrm{Interp}(\boldsymbol{x}_{\boldsymbol{r}}, \mathcal{T} ))$, 
where $\mathrm{Decoder}$ is an MLP with softplus activation, $\mathrm{Interp}$ denotes interpolation. The pixel value is calculated by volume rendering as:
\begin{equation}
    I(\mathbf{r})=\int_{s_{n}}^{s_{f}} p(s) \sigma(\mathbf{r}(s)) \mathbf{c}(\mathbf{r}(s)) \text{d}s,~\text{where}~ p(t)=\exp \left(-\int_{s_{n}}^{s} \sigma(\mathbf{r}(s)) \text{d}s\right)\text{.}
\end{equation}

We compute the non-saturating GAN loss~\citep{goodfellow2020generative} $\mathcal{L}_\text{img}$ and $\mathcal{L}_\text{vid}$ and R1 regularization loss~\citep{mescheder2018training} $L_{\text{R1}}$. Following EG3D, we use dual image discriminator which takes both the low-resolution raw image and high-resolution image as inputs. We also compute density regularization $L_{\sigma}$ on the generated video frames. The overall loss is formulated as:
\begin{equation}
\mathcal{L}_{\text{adv}}^{\mathcal{G}}=\mathcal{L}_\text{img}^{\mathcal{G}}+\mathcal{L}_{\text{vid}}^{\mathcal{G}}+\mathcal{L}_{\sigma}\text{,}~\mathcal{L}_{\text{adv}}^{\mathcal{D}}=\mathcal{L}_{\text{img}}^{\mathcal{D}}+\mathcal{L}_{\text{vid}}^{\mathcal{D}}+\mathcal{L}_{\text{R1}}.
\end{equation}

{\bf Inference.} 
Although trained on sparse frames only, our generator can synthesize contiguous frames during inference. For each video, we generate frames at timestep $t_i$, where $i \in \{0, 1, ..., N\}$, $N$ denotes the maximum number of frames. The mapping network in our generator takes camera pose $c_i$ for each frame to generate intermediate appearance code $\mathbf{w}_a^i$. However, during training, each frame has its camera pose. This brings discrepancy in intermediate appearance codes within the video, which harms the temporal consistency. We propose to share the same $c_i$ for the mapping network in inference, which largely improves temporal consistency as demonstrated in experiments.

%% file: Experiment.tex
\section{Experiments}
We study the following questions in our experiments. 1) Can \ours{} generate high-quality portrait videos with dynamic 3D geometry and multi-view consistency? 2) How does each component in our \ours{} model take effect? 3) What are the performance of \ours{} in downstream applications?
To answer these, we conduct extensive experiments on several human portrait video datasets.

\subsection{Datasets}
We experiment on three face video datasets, \ie, {\bf VoxCeleb}~\citep{nagrani2017voxceleb, chung2018voxceleb2}, {\bf CelebV-HQ}~\citep{zhu2022celebvhq}, and {\bf TalkingHead-1KH}~\citep{wang2021facevid2vid}. These datasets contain talking face clips of different identities extracted from online videos. We balance the video clips for each identity and preprocess the videos using a standard pipeline (Appendix~\ref{appendix:process}).

\subsection{Comparisons}
{\bf Baselines}.
We compare \ours{} against four state-of-the-art methods for 3D-aware video generation: (1) the concurrent work \textbf{3DVidGen}~\citep{bahmani20223d}; (2) \textbf{StyleNeRF+MCG-HD}: combining the SOTA 3D image GAN model StyleNeRF~\citep{gu2021stylenerf} with the SOTA multi-stage video generation work MoCoGAN-HD~\citep{tian2021good}; (3) \textbf{EG3D+MCG-HD}: combining EG3D~\citep{chan2022efficient} with MoCoGAN-HD; (4) \textbf{3DVidGen (EG3D)}: replacing the 3D image GAN backbone in 3DVidGen with EG3D.

{\bf Evaluation metrics}.
We evaluate \ours{} and baseline models  by Frechet Video Distance (FVD)~\citep{unterthiner2019fvd}, Multi-view Identity Consistency (ID)~\citep{shi2021lifting}, Chamfer Distance (CD), and Multi-view Image Warping Errors (WE)~\citep{zhang2022avatargen, zhang2022multi}. The evaluation metrics for multi-view consistency are originally proposed for 3D image generation. We extend these metrics to multiple frames which are suitable for 3D video generation tasks. Please refer to Appendix~\ref{appendix:metrics} for more details on our evaluation metrics.

\begin{figure}[t]
\centering
\begin{subfigure}[]{\textwidth}
\includegraphics[width=\textwidth]{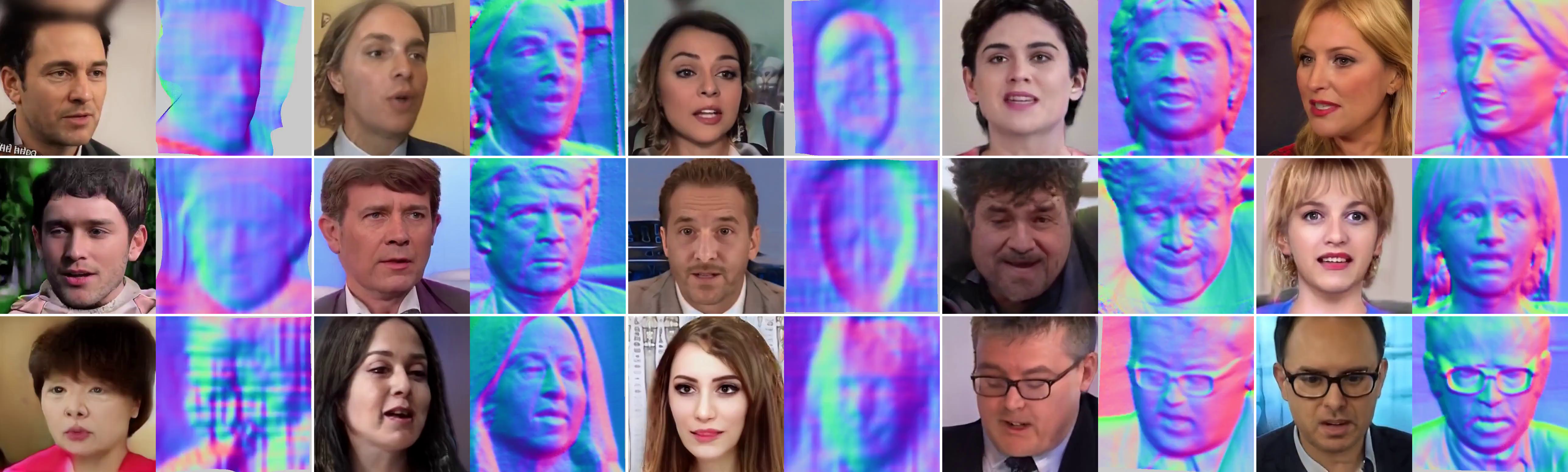}
\end{subfigure}
\begin{subfigure}[]{\textwidth}
\centering
\includegraphics[width=\textwidth]{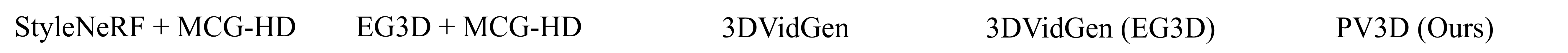}
\end{subfigure}
\vspace{-4mm}
\caption{Qualitative comparisons between \ours{} and baselines, 
see our \href{https://showlab.github.io/pv3d}{project page} for videos.}
\label{fig:qual}
\vspace{-4mm}
\end{figure}

{\bf Quantitative evaluations}.
Table~\ref{tab:comp} summarizes the quantitative comparisons between \ours{} and baseline models. First, we observe that \ours{} outperforms all of the baseline models on all datasets \textit{w.r.t.} FVD, ID, and WE, which shows our \ours{} can generate videos with diverse content and plausible motions while maintaining high multi-view consistency within videos.
It is worth noting that our \ours{} has a higher CD than StyleNeRF+MCG-HD. This is because CD computes the distance between two point clouds rendered from frontal and side views. As illustrated in Figure~\ref{fig:qual}, StyleNeRF+MCG-HD fails to synthesize detailed 3D geometry, thus leading to lower CD than \ours{}. A similar observation is made in TalkingHead-1KH where MCG-HD+EG3D baseline fails to generate videos with diverse motion, which also leads to smaller CD. On the other hand, \ours{} is able to generate high-quality {\it dynamic} 3D geometry with better multi-view consistency.

{\bf Qualitative evaluations}.
Figure~\ref{fig:qual} shows the qualitative comparisons.
We first observe that both StyleNeRF+MCG-HD and EG3D+MCG-HD produce poor results because their multi-stage framework design is not end-to-end trainable, resulting in implausible motions. 
{Compared with them, 3DVidGen and its  EG3D counterpart achieve relatively better video fidelity. However, the geometry quality and motion diversity of 3DVidGen are still not comparable with \ours{} due to their straightforward motion condition design.}
Differently, our \ours{} produces temporally consistent and motion plausible videos with high-quality geometry.

\begin{table}[ht]
\centering
  \vspace{-2.4mm}
  \caption{Quantitative comparisons, with best results bold and second best underlined.
  }
  \vspace{-1.6mm}
  \label{tab:comp}
  \renewcommand{\tabcolsep}{2.8pt}
  \resizebox{0.94\textwidth}{!}{
  \centering
  \begin{tabular}{@{\hskip 1mm}lcccc|cccc|cccc@{\hskip 1mm}}
    \toprule
    &\multicolumn{4}{c|}{VoxCeleb} &\multicolumn{4}{c|}{CelebV-HQ} &\multicolumn{4}{c}{TalkingHead-1KH}\\
    & FVD$\downarrow$  & ID$\uparrow$ & CD$\downarrow$ & WE$\downarrow$ &FVD$\downarrow$  & ID$\uparrow$ & CD$\downarrow$ & WE$\downarrow$ &FVD$\downarrow$  & ID$\uparrow$ & CD$\downarrow$ & WE$\downarrow$\\
    \midrule
    StyleNeRF+MCG-HD & 348.7      & 0.70       & {\bf 1.08} & 36.06   &134.4 &\underline{0.80} &{\bf 1.13} &38.73 & 292.7 &0.75 &5.34 &49.29 \\
    EG3D+MCG-HD      & 222.1      & \underline{0.80}       & 1.57       & \underline{10.57} &298.4 &0.77 &3.34 &\underline{10.74} &262.4 &\underline{0.78} &{\bf 1.39} & \underline{11.54}     \\
    3DVidGen         & 65.5       & 0.75       & 3.40       & 44.55   &\underline{63.6} &0.77 &3.80 &37.30 &\underline{83.0} &0.76 &4.35 &46.47 \\
    3DVidGen (EG3D)  & \underline{56.3}       & 0.71       & 3.65       & 24.55 &66.2 &0.70 &3.83 &26.34 &89.8 &0.65 &4.56 &35.48   \\
    \midrule
    \ours{} (Ours)     & {\bf 29.1} & {\bf 0.81} & \underline{1.34} & {\bf 9.76} & {\bf 39.3} &{\bf 0.81} &\underline{1.21} &{\bf 8.18} & {\bf 66.6} &{\bf 0.80} & \underline{2.33} &{\bf 10.73}\\
    \bottomrule
  \end{tabular}}
\vspace{-2mm}
\end{table}

\subsection{Ablation Studies}
\label{exp:ab}
In this section, we conduct ablation studies of \ours{} on VoxCeleb dataset as it contains more diverse motions and appearances as well as more balanced identity distributions. 

\begin{table}[ht]
\vspace{-4mm}
	\renewcommand{\tabcolsep}{0.8pt}
	\small
	\caption{Ablations of \ours{} on VoxCeleb. We vary the motion layers, motion generator architecture, camera conditions, and discriminator architectures to study their effects.
	}
	\begin{subtable}[!t]{0.24\textwidth}
		\centering
		\begin{tabular}{cccc}
			\toprule
			\textit{Pos.} & FVD$\downarrow$ & CD$\downarrow$ & WE$\downarrow$ \\
			\midrule
			  early  & 40.2 & 2.32 & 10.60\\
			  middle & 29.1 & 1.34 & 9.76\\
                late   & 33.3 & 6.24 & 11.56 \\
                learned  & 30.5 & 1.81 & 10.31\\
			\bottomrule
		\end{tabular}
		\caption{The effect of motion layer positions.}
		\label{tab:ablation:layers}
	\end{subtable}
	\hspace{\fill}
	\begin{subtable}[!t]{0.24\textwidth}
		\centering
		\begin{tabular}{cccc}
			\toprule
			\textit{Mot.} & FVD$\downarrow$ & CD$\downarrow$ & WE$\downarrow$ \\
			\midrule
                MLP  & 41.1 & 1.33 & 9.86\\
                Naive &36.9 &1.05 &10.21 \\
                $\mathbf{z}_m$$\rightarrow$$\mathbf{z}_a$  & 37.7 & 1.07 & 9.86 \\
                Ours & 29.1 & 1.34 & 9.76\\                
			\bottomrule
		\end{tabular}
		\caption{The effect of motion generator architectures.
		}
		\label{tab:ablation:arch}
	\end{subtable}
	\hspace{\fill}
 	\begin{subtable}[!t]{0.24\textwidth}
		\centering
		\begin{tabular}{cccc}
			\toprule
			\textit{Cam.} & FVD$\downarrow$ & CD$\downarrow$ & WE$\downarrow$ \\
			\midrule
			All  & 32.1 & 2.49 & 11.32\\
                Non & 55.3 & 3.28 & 23.71\\
			Map & 38.0 & 1.47 & 10.19\\
                MapT & 29.1 & 1.34 & 9.76\\
			\bottomrule
		\end{tabular}
		\caption{The effect of camera condition in generator.}
		\label{tab:ablation:cam}
	\end{subtable}
	\hspace{\fill}
 	\begin{subtable}[!t]{0.24\textwidth}
            \centering
		\begin{tabular}{cccc}   
			\toprule
			\textit{Vid. Dis.} & FVD$\downarrow$ & CD$\downarrow$ & WE$\downarrow$ \\
			\midrule
			w/o Cam & 34.9  &4.38  &10.84 \\
			w/o $\Delta t$  & 38.7  & 1.54 & 10.04 \\
		      w/ both& 29.1 & 1.34 & 9.76\\
			\bottomrule
		\end{tabular}
		\caption{The effect of different conditions in video discriminator.}
		\label{tab:ablation:discc}
	\end{subtable}
	\label{ablations}
	\vspace{-8mm}
\end{table}

{\bf Motion layer position.}
\ours{} computes and fuses motion features in selected synthesis layers, \ie, the first $K$ layers in our motion generator with a default setting of $K$ = 4 (denoted as middle). To study the effect of motion layer position, we insert motion layers until early ($K$ = 2) and late ($K$ = 7) stages. Moreover, we design a simple learnable parameter for each motion layer and optimize this weight for motion feature fusion (denoted as learned). The results are summarized in Table~\ref{tab:ablation:layers}. 

We observe that when the motion layers are too shallow, the model's capacity for modeling dynamics is limited and both video quality and multi-view consistency degrade. However, if we insert motion features into all the synthesis layers (late), FVD is still higher and 3D geometry is largely affected because (1) appearance feature manipulation space is large but hard to optimize, which harms motion plausibility; (2) our training process is highly under-constrained. In this case, the model can easily overfit the RGB frames, which hinders the learning of geometry. 
Similarly, although the video quality improves when using learnable fusion weights, the multi-view consistency is not as good as reflected by CD and WE metrics because the model still tends to produce better RGB content at the cost of worse geometry due to the lack of 3D supervision.

{\bf Motion generator.}
Table~\ref{tab:ablation:arch} summarizes the ablations on our motion generator architectures. Our video generator takes two random codes, \ie, $\mathbf{z}_a$ and $\mathbf{z}_m$. We first study the effect of introducing an independent motion code by replacing motion code $\mathbf{z}_m$ with $\mathbf{z}_a$. It can be observed that although CD reduces by 0.3, all other metrics deteriorate especially FVD, suggesting that motion plausibility is affected when motion code is removed. A similar observation is made in naive implementation.
Replacing modulated convolution with simple MLP significantly increases FVD, implying MLP is less effective than modulated convolution in manipulating appearance features.

{\bf Camera conditions.} 
We investigate how different camera conditioning strategies could affect generation performance. 
In each training iteration, we sample two video frames along with their camera poses. Thus, we have three options for camera conditioning (sharing camera pose or not) in our generator: (1) \textbf{All}: condition the whole generator \textcolor{black}{(both mapping network and rendering)} on the shared camera pose; (2) \textbf{Non}: condition the whole generator on the camera pose of each frame; (3) \textbf{Map}: only share camera pose for mapping network, and use different camera pose to render frames.

As shown in Table~\ref{tab:ablation:cam}, sharing one camera has poor multi-view consistency, with 2.49 CD and 11.32 WE. On the other hand, using non-shared cameras leads to even worse performance.
{This is because our mapping network takes a camera pose when computing appearance code. Therefore, changing the camera across video frames would bring rapid changes to the appearance code, leading to temporal inconsistency.}
{An alternative is to only share camera poses in the mapping network. However, this strategy works poorly because the camera pose discrepancy in generation process hinders the convergence. On the contrary, sharing the same camera pose in the mapping network only during inference stage (\textbf{MapT}) facilitates the training and preserves temporal consistency.} \textcolor{black}{Hence, we use \textbf{MapT} as our default setting.}

{\bf Discriminator conditions.} 
Table~\ref{tab:ablation:discc} shows the effect of camera pose and time difference conditions in the video discriminator. 
We can observe that camera conditions in video discriminator can largely improve 3D geometry, while time difference helps improve temporal coherence by encoding auxiliary temporal information.

\subsection{Applications}
\begin{figure}[h]
\vspace{-2mm}
\centering
\includegraphics[width=0.65\textwidth]{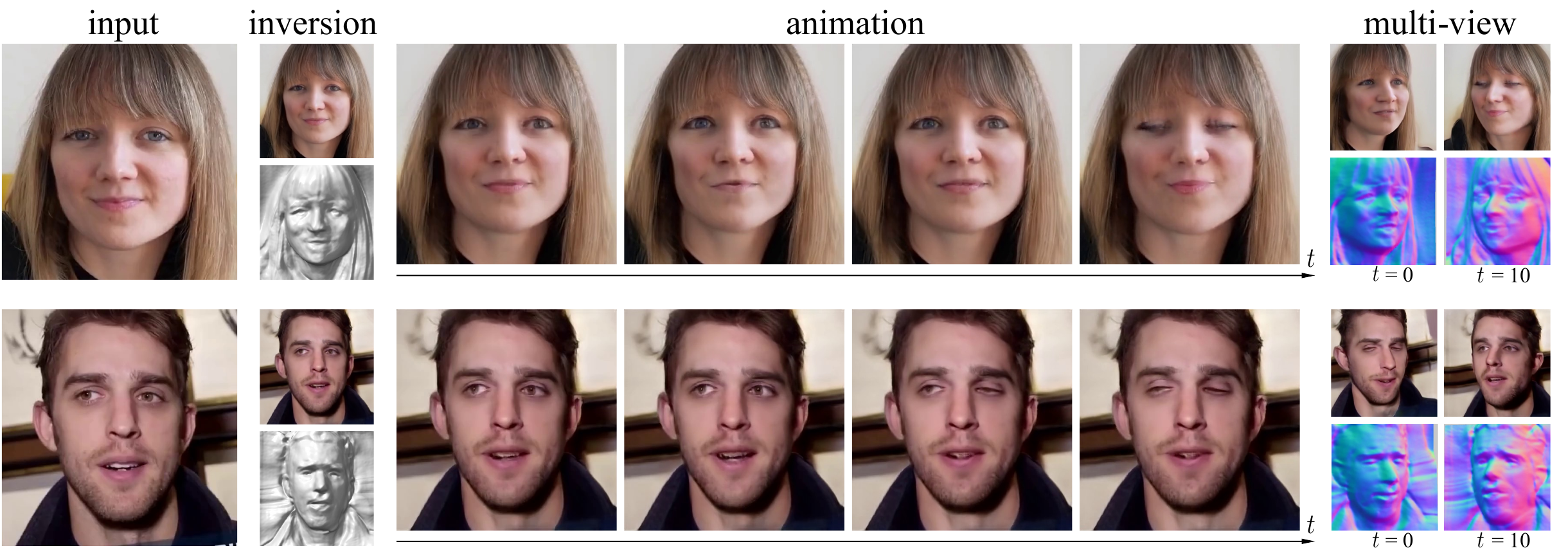}
\caption{Given static portraits, we fix the video generator and optimize the intermediate appearance code in $\mathcal{W}^+$ space to perform inversion. 
By sampling a random motion code, we can animate the static portraits with natural motion and synthesize portrait videos with multi-view consistency.}
\vspace{-2mm}
\label{fig:app1}
\end{figure}

{\bf Static portrait animation.} 
Our generator can independently generate video frame at a certain timestep instead of generating from the first frame auto-regressively. This flexible architecture enables static portrait animation. Given the input image and the estimated camera pose, we fix our generator and optimize latent code at timestep $t$ = 0. The inversion is performed in $\mathcal{W}^+$ space~\citep{richardson2021encoding,abdal2019image2stylegan}. As shown in Figure~\ref{fig:app1}, the GAN inversion based on our generator also produces high-quality 3D shape for the input frame. We then keep the latent code fixed and randomly sample a motion code to drive the portrait with natural motion. With the 3D priors learned by our \ours{}, the synthesized videos can also be rendered under arbitrary viewpoints.

\begin{figure}[h]
\centering
\vspace{-2mm}
\includegraphics[width=0.87\textwidth]{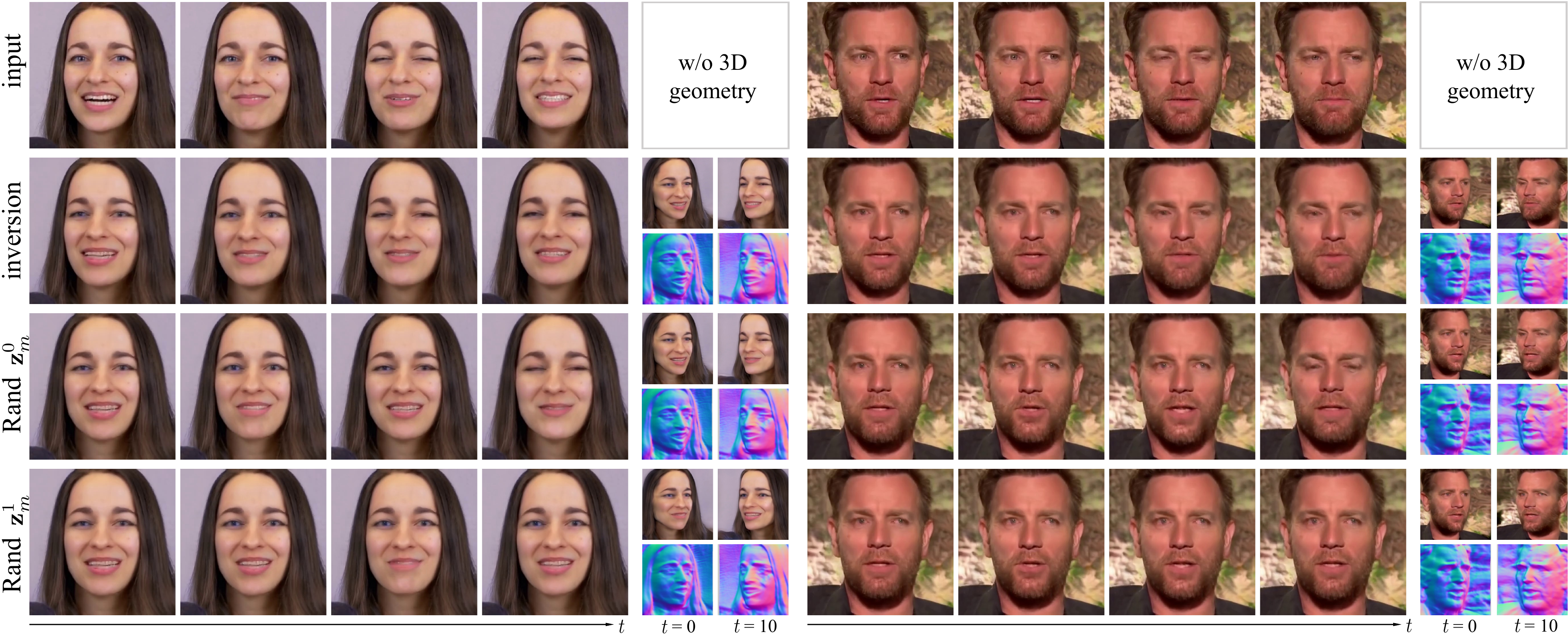}
\caption{\ours{} provides a quick solution for 3D reconstruction on monocular videos. Thanks to the disentanglement of appearance and motion in \ours{}, the motion of input videos can be manipulated by changing motion codes. The results can still maintain multi-view consistency.}
\vspace{-2mm}
\label{fig:app2}
\end{figure}

{\bf Monocular video reconstruction and motion editing.}
Given a video and its pre-estimated camera pose sequence, we can directly reconstruct the video based on our pretrained generator. For the video content, we also optimize the intermediate appearance code in $\mathcal{W}^+$ space. As for the motion component, we experimentally find that $\mathbf{z}_m^+$ space is more effective. Specifically, we inverse $\mathbf{z}_m$ for each video frame individually. Figure~\ref{fig:app2} illustrates the results for reconstruction, our \ours{} provides a simple solution for 3D reconstruction on monocular videos. Thanks to the disentangled design of motion and appearance components in \ours{}, we can fix appearance codes and sample new motion codes to manipulate the motion of input videos in the 3D domain.

%% file: Conclusion.tex
\section{Conclusion}
This work introduces the first 3D-aware generative model, \ours{}, for synthesizing multi-view consistent portrait videos with high-quality 3D geometry. By employing independent latent codes for appearance and motion, \ours{} can leverage temporal tri-plane synthesis to address the challenges in 3D-aware portrait video generation. Moreover, we condition \ours{} on camera pose sequence to alleviate the challenging motion ambiguities. We demonstrate that \ours{} can generate both temporal and multi-view consistent portrait videos with diverse motions and dynamic 3D geometry. Besides, \ours{} supports downstream applications such as static portrait animation, 3D video reconstruction, and multi-view consistent motion editing. We believe our method will facilitate the desired practical applications in VR/AR and visual effects.

%% file: Acknowledgements.tex
\section{Acknowledgements}
This project is supported by the National Research Foundation, Singapore under its NRFF Award NRF-NRFF13-2021-0008 and the Ministry of Education, Singapore, under the Academic Research Fund Tier 1 (FY2022).

%% file: Ethnics_statement.tex
\section{Ethnics Statement}
Static portrait animation, monocular video reconstruction, and editing could be misused for generating fake videos or manipulating authentic videos for improper or illegal purposes. These potentially harmful applications may pose a societal threat. We strictly forbid these kinds of abuses. In addition, our video generator may contain bias for face results due to the unbalanced video distribution in training datasets.

%% file: Reproducibility_statement.tex
\section{Reproducibility Statement}
To ensure reproducibility, we describe the implementation details in Appendix.~\ref{appendix:details}. The dataset preprocessing steps are introduced thoroughly in Appendix.~\ref{appendix:process}, including video sources, training data balance, face video alignment pipeline, and camera pose estimation approach.
Our code and models are publicly available at \href{https://showlab.github.io/pv3d}{https://showlab.github.io/pv3d}.

%% file: Appendix.tex
\section{Appendix}
\subsection{Implementation Details}
\label{appendix:details}
For each video, we sample two frames within a 16-frame span. Following DiGAN~\citep{yu2022generating}, we sample timesteps $\{t_i, t_j\}$ from beta distributions. The resolution of the generated video is 512$\times$512. We use a resolution of 64 and a sampling step of 48 for neural rendering during training. In inference stage, we use a rendering resolution of 128 for geometry visualization only. Each camera pose $c$ has 25 dimensions, with 16 for extrinsics and 9 for intrinsics. Our model is implemented using PyTorch. We balance the loss terms by weighting factors: 1) $\lambda_{\text{reg}}$=\text{0.6}, $\lambda_{\text{vid}}$=\text{0.65}, $\lambda_{\text{img}}$=\text{1.0}, $\lambda_{\text{R}_1}$=\text{2.0} for VoxCeleb; 2) $\lambda_{\text{reg}}$=\text{0.05}, $\lambda_{\text{vid}}$=\text{0.65}, $\lambda_{\text{img}}$=\text{1.0}, $\lambda_{\text{R}_1}$=\text{4.0} for CelebV-HQ; 3) $\lambda_{\text{reg}}$=\text{0.5}, $\lambda_{\text{vid}}$=\text{0.65}, $\lambda_{\text{img}}$=\text{1.0}, $\lambda_{\text{R}_1}$=\text{2.0} for TalkingHead-1KH.
Our model is trained for 300k iterations with a batch size of 16, which takes 58 hours on 8 Nvidia A100 GPUs.

\subsection{Analysis of Latent Code Space}
\label{appendix:wanalysis}
The architecture of synthesis layer in our \ours{} largely follows StyleGAN and its following works~\citep{karras2019style, karras2020analyzing, karras2021alias}. Based on the pre-trained StyleGAN models, prior works~\citep{shen2020interpreting} also investigate the property of the latent code space. These works show that the intermediate latent code space has extensive manipulation ability for image synthesis. Although the original design of StyleGAN is sharing one intermediate latent code across all synthesis layers, follow-up works~\citep{abdal2019image2stylegan,richardson2021encoding,tov2021designing} relax this constraint and achieve better reconstruction results for image inversion because using different latent codes for different synthesis layers can further expand the space for image generation. 
\begin{figure}[!h]
\centering
\includegraphics[width=0.85\textwidth]{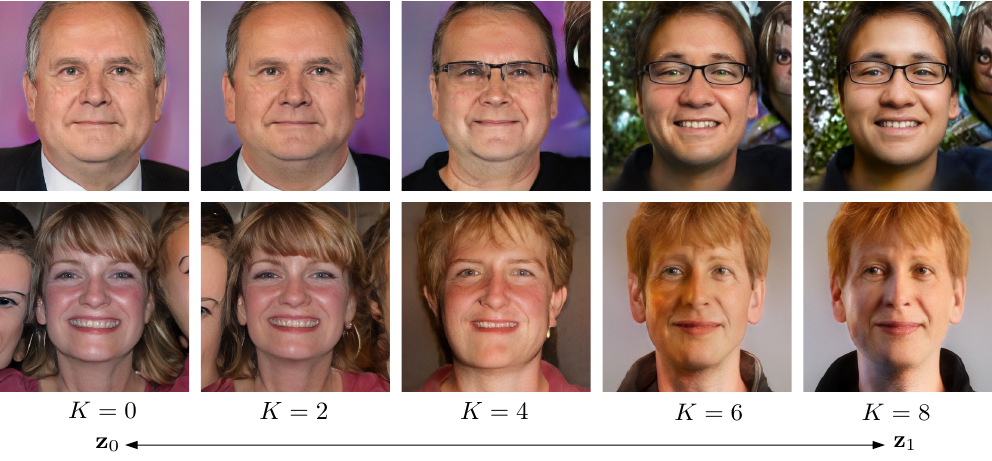}
\caption{Results of style mixing in first $K$ synthesis layers in pre-trained EG3D. For each example, we first sample one latent code and mix it with another latent code in the first $K$ synthesis layers.}
\label{fig:wanalysis}
\end{figure}

Our \ours{} can generate temporal tri-plane features by modulating appearance features based on motion code and timestep. We encode motion code and timestep into intermediate motion codes and then compute motion features in temporal tri-plane synthesis network. Because our synthesis network is built on top of EG3D, we also analyze the latent space to find out how the manipulation of appearance features could affect the synthesis results. As shown in Figure~\ref{fig:wanalysis}, we perform style mixing in the first $K$ synthesis layers of a pre-trained EG3D model. When $K$ increases, the image contents gradually change. Specifically, manipulating the appearance code in $K$ = 2 layers can largely preserve the contents. However, only modulating features in the first 2 layers would potentially harm the capacity for content diversity of our video generator. When $K$ $\ge$ 6, there exists a sharp change in the image content. Because one important property for portrait video is the temporal coherence, \ie, consistent identity, we finally select $K$ = 4 in our motion generator to maintain a good temporal consistency as well as motion diversity.

\subsection{Dataset Preprocessing}
\label{appendix:process}
{\bf Video sources.} 

{\it VoxCeleb}~\citep{nagrani2017voxceleb, chung2018voxceleb2} is an audio-visual speaker verification dataset containing interview videos for more than 7,000 speakers. It provides speaker labels for each video clip. For each speaker, we sample two video clips that have the highest video resolutions.

{\it CelebV-HQ}~\citep{zhu2022celebvhq} is a large-scale face video dataset that provides high-quality video clips involving 15,653 identities. Compared with VoxCeleb, it contains diverse lighting conditions.

{\it TalkingHead-1KH}~\citep{wang2021facevid2vid} consists of talking head videos extracted from 2,900 long video conferences.

For all of the datasets, we directly download videos with the highest possible resolution from YouTube using the provided uid list. 

{\bf Training data balance.} CelebV-HQ and TalkingHead-1KH have unbalanced number of video clips for each identity. To balance the video clips, we perform face clustering to predict pseudo-identities and sample video clips for each identity. Specifically, we extract facial features using pre-trained ArcFace~\citep{deng2019arcface} and employ agglomerative hierarchical clustering~\citep{day1984efficient} to predict pseudo-identities and sample at most two videos for each identity.

{\bf Alignment.} All datasets are preprocessed using the same alignment pipeline:
We first detect landmarks for each frame using an off-the-shelf face alignment package~\citep{bulat2017far}.
Such single frame face alignment technique would introduce temporal inconsistency. Following \citep{fox2021stylevideogan}, we use a low-pass Gaussian filter to smooth the estimated keypoints before warping the images. Then, we follow the image warping approach of FFHQ~\citep{karras2019style} to align each frame. To facilitate the training of unconditional 3D image GAN, EG3D employs an extra cropping step to process the image. Specifically, they realign the image in depth direction, which forces all of the keypoint of nose to the same point in the world coordinate defined by the parametric face model, \ie, 3DMM~\citep{bfm09}. We follow this step to process the video clips.
Finally, we apply deep face reconstruction~\citep{deng2019accurate} to estimate camera pose for each video frame. Again, this process brings temporal inconsistency and we also apply the low-pass Gaussian filter to smooth the results.

{\bf Verification.} Our preprocessing pipeline is automated and purely based on off-the-shelf packages. However, there exist noise and failure cases in each preprocessing step. Therefore, we apply an extra verification step to remove the noisy video clips. In particular, we use ArcFace to extract the facial features again for every 2 frames within a video clip. If the similarity scores between one frame and others are below a threshold $\tau=0.5$, this frame will be labeled as noisy. We discard video clips that contain more than two noisy frames.

\subsection{Evaluation Metrics}
\label{appendix:metrics}
{\bf Frechet Video Distance (FVD).} We compute the statistics for ground-truth and generated samples using the pre-trained I3D~\citep{carreira2017quo} model (the PyTorch version checkpoint released by VideoGPT\footnote{https://github.com/wilson1yan/VideoGPT}). For the training dataset, we randomly sample 5000 videos. Each video generator synthesizes 5000 uncurated videos randomly for test. All of the videos are encoded in H264 format. Compared with previous methods that save videos in PNG format~\citep{skorokhodov2021stylegan}, we empirically find that saving videos with H264 and decoding videos into images during testing does not cause any variation and can largely save space.

{\bf Multi-view Identity Consistency (ID).} 3D-aware image GAN works compute ID by rendering both frontal view and side view images to measure the model's multi-view identity consistency. They adopt state-of-the-art face recognition model~\citep{deng2019arcface} (denoted as $\mathcal{F}$) to extract features and compute similarity scores between frontal and side view images. We simply extend this process to multiple frames. In our experiments, we generate random videos and render both frontal face and side face images $I_{\text{y}}$, at two randomly sampled timesteps $\{t_\text{0}$, $t_\text{1}\}$, where y denotes yaw angles. The ID metric is formulated as:
\begin{equation}
    \text{ID}(I_{\text{y}_\text{0}}^{t_\text{0}}, I_{\text{y}_\text{1}}^{t_\text{1}}) = \mathcal{F}(I_{\text{y}_\text{0}}^{t_\text{0}})^{T}\mathcal{F}(I_{\text{y}_\text{1}}^{t_\text{1}})\text{,}
\end{equation}
where $t_\text{0}\ne t_\text{1}$ or $\text{y}_\text{0} \ne \text{y}_\text{1}$. In this work, we report results with y $\in\{\text{0}^{\circ}, \text{30}^{\circ}\}$ on 1000 videos. We compute ID for each frame pair within each video. The final ID metric is the average value for all of the image pairs.

{\bf Chamfer Distance (CD).} StyleSDF~\citep{or2021stylesdf} proposes to use the Chamfer Distance between the frontal and side view point clouds to measure the multi-view consistency of 3D geometry. In this work, we extend this metric to multiple frames by computing CD between depth map pairs in each video. For each video, we sample two timesteps $\{t_\text{0}$, $t_\text{1}\}$ and render point cloud at two angles $P_{\text{y}}$, where y denotes yaw angles. The CD is mathematically formulated as:
\begin{equation}
\text{CD}\left(P_{\text{y}_\text{0}}^{t_\text{0}}, P_{\text{y}_\text{1}}^{t_\text{1}}\right)=\underset{{x \in P_{\text{y}_\text{0}}^{t_\text{0}}}}{\text{med}} \min_{y \in P_{\text{y}_\text{1}}^{t_\text{1}}}\|x-y\|_2^2+\underset{y \in P_{\text{y}_\text{1}}^{t_\text{1}}}{\text{med}} \min _{x \in P_{\text{y}_\text{0}}^{t_\text{0}}}\|x-y\|_2^2\text{,}
\end{equation}
where $\text{med}$ means median, $t_\text{0}\ne t_\text{1}$ or $\text{y}_\text{0} \ne \text{y}_\text{1}$. Following StyleSDF we normalize the point clouds based on the volume sampling bin size for each generator before computing CD. We also remove the non-terminating rays whose opacity is below 0.5 for all of the models whose backbone is EG3D. To make fair comparisons, for baselines based on StyleNeRF, we render point clouds in foreground NeRF, \ie, rendering face part only. We interpolate all of the points clouds to 64$\times$64. We also use y $\in\{\text{0}^{\circ}, \text{30}^{\circ}\}$ and render 1000 videos. We then average CD for all of the point cloud pairs to get the final CD result.

{\bf Multi-view Image Warping Errors (WE).} 
Inspired by recent 3D image GAN works~\citep{zhang2022multi,zhang2022avatargen}, we can further compute the warping error by reprojecting each pixel from side view to the frontal view based on the images $(I_{\text{y}_\text{0}}^{t_\text{0}}, I_{\text{y}_\text{1}}^{t_\text{1}}) $ and depth maps $(D_{\text{y}_\text{0}}^{t_\text{0}}, D_{\text{y}_\text{1}}^{t_\text{1}}) $ at yaw angle $\{\text{y}_\text{0}, \text{y}_\text{1}\}$ and timesteps $\{{t}_\text{0}, {t}_\text{1}\}$. The camera extrinsics and intrinsics are $[$R$|$t$]$ and K, where R is determined by yaw angle y. We warp a pixel located at $(i, j)$ (denoted as $\mathbf{x}$) in side view image to frontal view by re-projection. It is formulated as:
\begin{equation}
    \mathbf{x}'=\text{K}[\text{R}'|\text{t}'][\text{R}|\text{t}]^{-1}[\text{K}^{-1}\mathbf{x},D(\mathbf{x})]^T\text{.}
\end{equation}
Based on the re-projected coordinates, we compute the warping error as:
\begin{equation}
\text{WE}(I_{\text{y}_\text{0}}^{t_\text{0}}, I_{\text{y}_\text{1}}^{t_\text{1}})=\frac{1}{L}\sum|I_{\text{y}_\text{0}}^{t_\text{0}}-I_{\text{warp}}|,
\end{equation}
where $I_{\text{warp}}(\mathbf{x}')=I_{\text{y}_\text{1}}(\mathbf{x})$ and $\text{y}_\text{0} \ne \text{y}_\text{1}$, $L$ is the number of re-projected points that located within the field-of-view in the frontal view image, and we only compute WE based on these visible pixels. Both image and point cloud are resized to 256$\times$256. Finally, we use y $\in\{\text{0}^{\circ}, \text{30}^{\circ}\}$ and generate 1000 videos. We compute the warping error for each front and side view pair in one video. The WE reported in this work is the average value of all the pairs.

\subsection{Additional Experiment Results}
\label{appendix:exp}

{\bf Ablations.} We perform additional ablation study on \ours{} by varying the design of {\it discriminator components} and {\it volume rendering resolutions}. 
\begin{table}[ht]
	\renewcommand{\tabcolsep}{2pt}
	\small
	\caption{Ablations on discriminators, resolution for neural rendering, and generator camera pose conditions.
	}
 	\begin{subtable}[!t]{0.32\textwidth}
		\centering
		\begin{tabular}{cccc}
			\toprule
			\textit{Disc.} & FVD$\downarrow$ & CD$\downarrow$ & WE$\downarrow$ \\
			\midrule
			w/o Vid  & 145.9 & 3.89 & 11.82\\
			w/o Img & 42.2 & 3.44 & 14.40\\
            w/ both disc. & 29.1 & 1.34 & 9.76\\
			\bottomrule
		\end{tabular}
		\caption{The effect of discriminator components.}
		\label{tab:appendix:ab:discw}
	\end{subtable}
	\hspace{\fill}
 	\begin{subtable}[!t]{0.32\textwidth}
		\centering
		\begin{tabular}{cccc}
			\toprule
			\textit{Rend. Res.} & FVD$\downarrow$ & CD$\downarrow$ & WE$\downarrow$ \\
			\midrule
			32  & 42.4 & 3.25 & 12.98 \\
			64  & 29.1 & 1.34 & 9.76\\
			128 & 34.1 & 0.92 & 9.62\\
			\bottomrule
		\end{tabular}
		\caption{The effect of volume rendering resolution.}
		\label{tab:appendix:ab:res}
	\end{subtable}
        \hspace{\fill}
        \textcolor{black}{
 	\begin{subtable}[!t]{0.32\textwidth}
		\centering
		\begin{tabular}{cccc}
			\toprule
			\textit{Gen. Cam.} & FVD$\downarrow$ & CD$\downarrow$ & WE$\downarrow$ \\
			\midrule
                w/o cam  & 44.3 & 2.35 & 13.01\\
                w/ cam & 29.1 & 1.34 & 9.76\\
			\bottomrule
		\end{tabular}
		\caption{\textcolor{black}{The effect of camera conditioning in generator.}}
		\label{tab:appendix:ab:cam}
        \end{subtable}
        }
	\label{ablations}
	\vspace{-2mm}
\end{table}

{\it Discriminator components.} \ours{} employs two independent discriminators to regularize the video content and motions. We ablate the image and video discriminators to study how each one supervises the training of generator. Table~\ref{tab:appendix:ab:discw} illustrates the results of removing each discriminator. Without video discriminator, the video quality largely deteriorates, suggesting that video discriminator is able to guarantee motion plausibility. In addition, removing image discriminator also brings a significant performance drop in FVD, CD, and WE, which proves that our video discriminator can also supervise the frame content but struggle to guarantee 3D geometry.

{\it Volume rendering resolution.} Table~\ref{tab:appendix:ab:res} summarizes the results of using different resolutions for neural volume rendering during training. A small resolution (32) restricts model's capacity, leading to worse performance on all of the metrics. Although a higher resolution (128) outperforms our default setting (64) in terms of multi-view consistency and warping error, its video quality is still worse. The results show that there exists a tradeoff between the video quality and 3D geometry in our generator because it is trained only on 2D videos. Considering the computation resources, our default setting uses a resolution of 64. In this work, we also use 64 for testing and reporting evaluation metrics. Moreover, we only use a resolution of 128 for geometry visualization.

\textcolor{black}{{\it Generator camera conditioning.} \ours{} takes camera poses to encode 3D priors in generator. Table~\ref{tab:appendix:ab:cam} shows that the 3D priors improve performance with a large margin. Without the camera pose condition in generator, a large decrease can be observed in all of the evaluation metrics.}

\begin{figure}[t]
\centering
\begin{subfigure}[]{0.49\textwidth}
\centering
\includegraphics[width=0.89\textwidth]{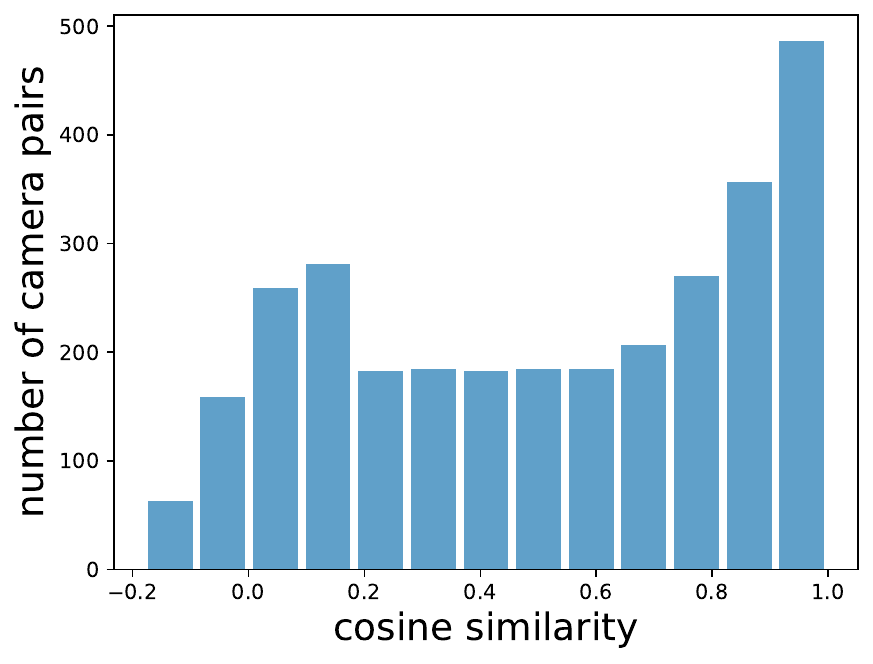}
\caption{\textcolor{black}{Statistics for cosine similarities between the embeddings of original and flipped camera pose pairs.}}
\label{fig:appendix:exp:disc:a}
\end{subfigure}
\hfill
\begin{subfigure}[]{0.5\textwidth}
\centering
\includegraphics[width=0.9\textwidth]{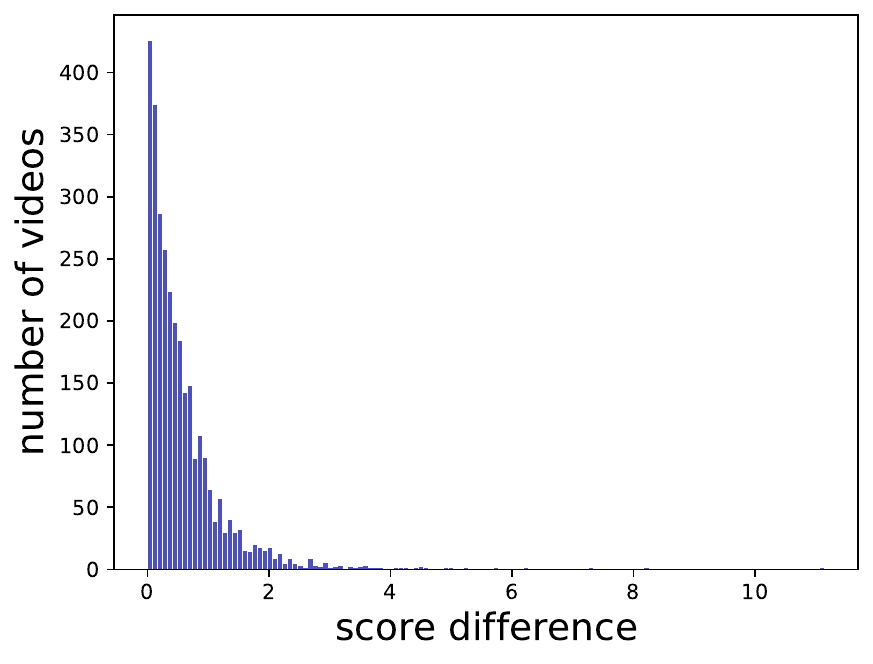}
\caption{\textcolor{black}{Statistics for real/fake score differences between original and flipped camera pose pairs.}}
\label{fig:appendix:exp:disc:b}
\end{subfigure}
\caption{\textcolor{black}{The analysis for the alignment ability of our video discriminator. We sample 3000 camera pose sequences from training dataset and input both original camera pose sequences together with the reversed ones into our video discriminator to study its ability for encoding ordinal information.}}
\label{fig:appendix:exp:disc}
\vspace{-4mm}
\end{figure}
\textcolor{black}{{\bf Video discriminator alignment.} To investigate whether our video discriminator can align the camera poses with the input video frames, we conduct two following experiments.}

\textcolor{black}{{\it Effects of camera pose order on embedding.} We sample 3000 camera pairs from the training dataset and pass them to our video discriminator. We then exchange the order of the camera poses and compute the embeddings again. After that, we compute the cosine similarities between the embeddings of original and the flipped camera pose sequences. The statistics for similarities are shown in Figure~\ref{fig:appendix:exp:disc:a}. It can be observed that most of the camera pose sequences with reversed orders will be mapped into embeddings that are far apart. Thus, our video discriminator is sensitive to the camera pose orders.}

\textcolor{black}{{\it Effects of camera pose order on discrimination ability.} We further input the videos generated by the sampled camera poses together with both original and flipped camera pose sequences into the video discriminator. We compute the L1 distances between the real/fake scores and visualize statistics in Figure~\ref{fig:appendix:exp:disc:b}. It illustrates that our video discriminator predicts different real/fake scores for most of the video frame pairs if the camera pose sequences are in reverse order. The results prove that our video discriminator is order-aware and can align the camera pose sequence with the video frame pair.}

{\bf Longer-term video synthesis.}
\ours{} is trained on two video frames per clip, this architecture makes it suitable for long-term video synthesis. To study its ability for long-term video generation, we further train and test \ours{} on video clips of 48 frames. The results are shown on our \href{https://showlab.github.io/pv3d}{project page}. As we can see, training data with longer duration contains more diverse motions and \ours{} can learn to generate such motions accordingly.

\begin{figure}[t]
\centering
\begin{subfigure}[]{0.9\textwidth}
\centering
\includegraphics[width=0.9\textwidth]{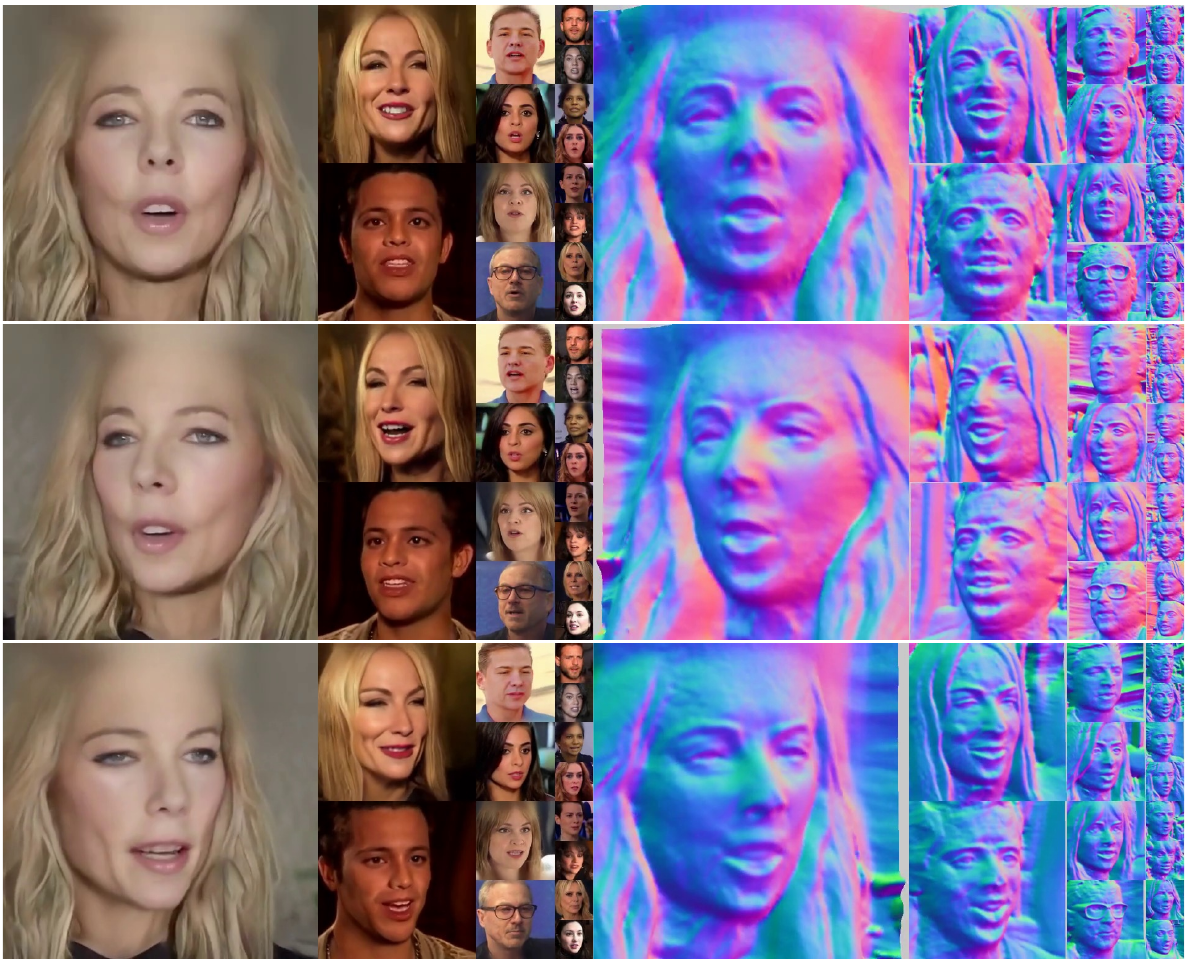}
\end{subfigure}

\begin{subfigure}[]{0.9\textwidth}
\centering
\includegraphics[width=0.9\textwidth]{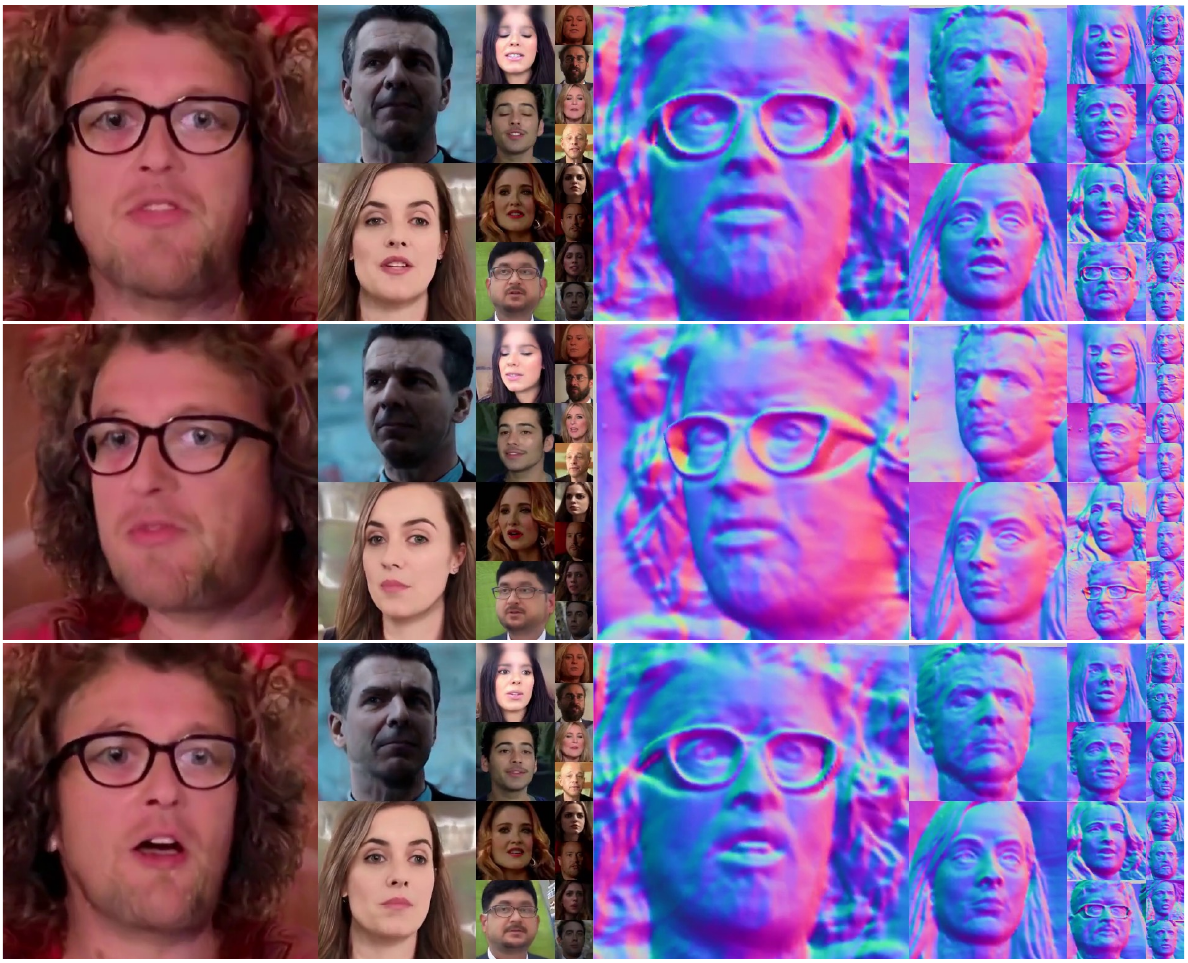}
\end{subfigure}
\caption{Samples generated by \ours{}, see our \href{https://showlab.github.io/pv3d/}{project page} for video results.
}
\label{fig:appendix:viz}
\vspace{-4mm}
\end{figure}

{\bf Qualitative results.}
We demonstrate more results of \ours{} in Figure~\ref{fig:appendix:viz}. 

\subsection{Limitations and Future Work}
Our \ours{} has several limitations: 1) \ours{} is trained and tested on video clips that contain at most 48 frames. The model's ability for modeling long-term (minutes order) dynamics is unknown. 2) The 2D video dataset quality is not comparable to image datasets such as FFHQ and CelebA. Our model has a flexible architecture that may support pre-training or joint training on image datasets, yet this augmentation strategy has not been explored although it is promising and meaningful. 
For future work, we will explore modeling long-term dynamics with novel 3D representations that are more suitable for 3D video generation, and leverage high-quality image datasets for data augmentation.

%% file: main.bbl
\begin{thebibliography}{56}
\providecommand{\natexlab}[1]{#1}
\providecommand{\url}[1]{\texttt{#1}}
\expandafter\ifx\csname urlstyle\endcsname\relax
  \providecommand{\doi}[1]{doi: #1}\else
  \providecommand{\doi}{doi: \begingroup \urlstyle{rm}\Url}\fi

\bibitem[Abdal et~al.(2019)Abdal, Qin, and Wonka]{abdal2019image2stylegan}
Rameen Abdal, Yipeng Qin, and Peter Wonka.
\newblock Image2stylegan: How to embed images into the stylegan latent space?
\newblock In \emph{{ICCV}}, 2019.

\bibitem[Abdal et~al.(2022)Abdal, Zhu, Mitra, and
  Wonka]{abdal2022video2stylegan}
Rameen Abdal, Peihao Zhu, Niloy~J Mitra, and Peter Wonka.
\newblock Video2stylegan: Disentangling local and global variations in a video.
\newblock \emph{arXiv preprint arXiv:2205.13996}, 2022.

\bibitem[Bahmani et~al.(2022)Bahmani, Park, Paschalidou, Tang, Wetzstein,
  Guibas, Van~Gool, and Timofte]{bahmani20223d}
Sherwin Bahmani, Jeong~Joon Park, Despoina Paschalidou, Hao Tang, Gordon
  Wetzstein, Leonidas Guibas, Luc Van~Gool, and Radu Timofte.
\newblock 3d-aware video generation.
\newblock \emph{arXiv preprint arXiv:2206.14797}, 2022.

\bibitem[Bulat \& Tzimiropoulos(2017)Bulat and Tzimiropoulos]{bulat2017far}
Adrian Bulat and Georgios Tzimiropoulos.
\newblock How far are we from solving the 2d \& 3d face alignment problem? (and
  a dataset of 230,000 3d facial landmarks).
\newblock In \emph{{ICCV}}, 2017.

\bibitem[Cao et~al.(2022)Cao, Simon, Kim, Schwartz, Zollhoefer, Saito,
  Lombardi, Wei, Belko, Yu, et~al.]{cao2022authentic}
Chen Cao, Tomas Simon, Jin~Kyu Kim, Gabe Schwartz, Michael Zollhoefer,
  Shun-Suke Saito, Stephen Lombardi, Shih-En Wei, Danielle Belko, Shoou-I Yu,
  et~al.
\newblock Authentic volumetric avatars from a phone scan.
\newblock \emph{{ACM Trans. on Graphics}}, 2022.

\bibitem[Carreira \& Zisserman(2017)Carreira and Zisserman]{carreira2017quo}
Joao Carreira and Andrew Zisserman.
\newblock Quo vadis, action recognition? a new model and the kinetics dataset.
\newblock In \emph{{CVPR}}, 2017.

\bibitem[Chan et~al.(2021)Chan, Monteiro, Kellnhofer, Wu, and
  Wetzstein]{chan2021pi}
Eric~R Chan, Marco Monteiro, Petr Kellnhofer, Jiajun Wu, and Gordon Wetzstein.
\newblock pi-gan: Periodic implicit generative adversarial networks for
  3d-aware image synthesis.
\newblock In \emph{{CVPR}}, 2021.

\bibitem[Chan et~al.(2022)Chan, Lin, Chan, Nagano, Pan, De~Mello, Gallo,
  Guibas, Tremblay, Khamis, et~al.]{chan2022efficient}
Eric~R Chan, Connor~Z Lin, Matthew~A Chan, Koki Nagano, Boxiao Pan, Shalini
  De~Mello, Orazio Gallo, Leonidas~J Guibas, Jonathan Tremblay, Sameh Khamis,
  et~al.
\newblock Efficient geometry-aware 3d generative adversarial networks.
\newblock In \emph{{CVPR}}, 2022.

\bibitem[Chung et~al.(2018)Chung, Nagrani, and Zisserman]{chung2018voxceleb2}
Joon~Son Chung, Arsha Nagrani, and Andrew Zisserman.
\newblock Voxceleb2: Deep speaker recognition.
\newblock \emph{INTERSPEECH}, 2018.

\bibitem[Day \& Edelsbrunner(1984)Day and Edelsbrunner]{day1984efficient}
William~HE Day and Herbert Edelsbrunner.
\newblock Efficient algorithms for agglomerative hierarchical clustering
  methods.
\newblock \emph{Journal of classification}, 1984.

\bibitem[Deng et~al.(2019{\natexlab{a}})Deng, Guo, Xue, and
  Zafeiriou]{deng2019arcface}
Jiankang Deng, Jia Guo, Niannan Xue, and Stefanos Zafeiriou.
\newblock Arcface: Additive angular margin loss for deep face recognition.
\newblock In \emph{{CVPR}}, 2019{\natexlab{a}}.

\bibitem[Deng et~al.(2019{\natexlab{b}})Deng, Yang, Xu, Chen, Jia, and
  Tong]{deng2019accurate}
Yu~Deng, Jiaolong Yang, Sicheng Xu, Dong Chen, Yunde Jia, and Xin Tong.
\newblock Accurate 3d face reconstruction with weakly-supervised learning: From
  single image to image set.
\newblock In \emph{{CVPRw}}, 2019{\natexlab{b}}.

\bibitem[Deng et~al.(2022)Deng, Yang, Xiang, and Tong]{deng2021gram}
Yu~Deng, Jiaolong Yang, Jianfeng Xiang, and Xin Tong.
\newblock Gram: Generative radiance manifolds for 3d-aware image generation.
\newblock \emph{{CVPR}}, 2022.

\bibitem[Doukas et~al.(2021)Doukas, Zafeiriou, and
  Sharmanska]{doukas2021headgan}
Michail~Christos Doukas, Stefanos Zafeiriou, and Viktoriia Sharmanska.
\newblock Headgan: One-shot neural head synthesis and editing.
\newblock In \emph{{ICCV}}, 2021.

\bibitem[Fox et~al.(2021)Fox, Tewari, Elgharib, and
  Theobalt]{fox2021stylevideogan}
Gereon Fox, Ayush Tewari, Mohamed Elgharib, and Christian Theobalt.
\newblock Stylevideogan: A temporal generative model using a pretrained
  stylegan.
\newblock \emph{arXiv preprint arXiv:2107.07224}, 2021.

\bibitem[Goodfellow et~al.(2020)Goodfellow, Pouget-Abadie, Mirza, Xu,
  Warde-Farley, Ozair, Courville, and Bengio]{goodfellow2020generative}
Ian Goodfellow, Jean Pouget-Abadie, Mehdi Mirza, Bing Xu, David Warde-Farley,
  Sherjil Ozair, Aaron Courville, and Yoshua Bengio.
\newblock Generative adversarial networks.
\newblock \emph{Communications of the ACM}, 2020.

\bibitem[Grassal et~al.(2022)Grassal, Prinzler, Leistner, Rother, Nie{\ss}ner,
  and Thies]{grassal2022neural}
Philip-William Grassal, Malte Prinzler, Titus Leistner, Carsten Rother,
  Matthias Nie{\ss}ner, and Justus Thies.
\newblock Neural head avatars from monocular rgb videos.
\newblock In \emph{{CVPR}}, 2022.

\bibitem[Gu et~al.(2022)Gu, Liu, Wang, and Theobalt]{gu2021stylenerf}
Jiatao Gu, Lingjie Liu, Peng Wang, and Christian Theobalt.
\newblock Stylenerf: A style-based 3d aware generator for high-resolution image
  synthesis.
\newblock In \emph{{ICLR}}, 2022.

\bibitem[Karras et~al.(2019)Karras, Laine, and Aila]{karras2019style}
Tero Karras, Samuli Laine, and Timo Aila.
\newblock A style-based generator architecture for generative adversarial
  networks.
\newblock In \emph{{CVPR}}, 2019.

\bibitem[Karras et~al.(2020)Karras, Laine, Aittala, Hellsten, Lehtinen, and
  Aila]{karras2020analyzing}
Tero Karras, Samuli Laine, Miika Aittala, Janne Hellsten, Jaakko Lehtinen, and
  Timo Aila.
\newblock Analyzing and improving the image quality of stylegan.
\newblock In \emph{{CVPR}}, 2020.

\bibitem[Karras et~al.(2021)Karras, Aittala, Laine, H{\"a}rk{\"o}nen, Hellsten,
  Lehtinen, and Aila]{karras2021alias}
Tero Karras, Miika Aittala, Samuli Laine, Erik H{\"a}rk{\"o}nen, Janne
  Hellsten, Jaakko Lehtinen, and Timo Aila.
\newblock Alias-free generative adversarial networks.
\newblock \emph{{NeurIPS}}, 2021.

\bibitem[Kato et~al.(2018)Kato, Ushiku, and Harada]{kato2018neural}
Hiroharu Kato, Yoshitaka Ushiku, and Tatsuya Harada.
\newblock Neural 3d mesh renderer.
\newblock In \emph{{CVPR}}, 2018.

\bibitem[Ma et~al.(2021)Ma, Simon, Saragih, Wang, Li, De~La~Torre, and
  Sheikh]{ma2021pixel}
Shugao Ma, Tomas Simon, Jason Saragih, Dawei Wang, Yuecheng Li, Fernando
  De~La~Torre, and Yaser Sheikh.
\newblock Pixel codec avatars.
\newblock In \emph{{CVPR}}, 2021.

\bibitem[Max(1995)]{max1995optical}
Nelson Max.
\newblock Optical models for direct volume rendering.
\newblock \emph{{IEEE Transactions on Visualization and Computer Graphics}},
  1995.

\bibitem[Mescheder et~al.(2018)Mescheder, Geiger, and
  Nowozin]{mescheder2018training}
Lars Mescheder, Andreas Geiger, and Sebastian Nowozin.
\newblock Which training methods for gans do actually converge?
\newblock In \emph{{ICML}}, 2018.

\bibitem[Mildenhall et~al.(2020)Mildenhall, Srinivasan, Tancik, Barron,
  Ramamoorthi, and Ng]{mildenhall2020nerf}
Ben Mildenhall, Pratul~P Srinivasan, Matthew Tancik, Jonathan~T Barron, Ravi
  Ramamoorthi, and Ren Ng.
\newblock Nerf: Representing scenes as neural radiance fields for view
  synthesis.
\newblock In \emph{{ECCV}}, 2020.

\bibitem[Nagrani et~al.(2017)Nagrani, Chung, and
  Zisserman]{nagrani2017voxceleb}
Arsha Nagrani, Joon~Son Chung, and Andrew Zisserman.
\newblock Voxceleb: a large-scale speaker identification dataset.
\newblock \emph{INTERSPEECH}, 2017.

\bibitem[Niemeyer \& Geiger(2021)Niemeyer and Geiger]{niemeyer2021giraffe}
Michael Niemeyer and Andreas Geiger.
\newblock Giraffe: Representing scenes as compositional generative neural
  feature fields.
\newblock In \emph{{CVPR}}, 2021.

\bibitem[Or-El et~al.(2022)Or-El, Luo, Shan, Shechtman, Park, and
  Kemelmacher-Shlizerman]{or2021stylesdf}
Roy Or-El, Xuan Luo, Mengyi Shan, Eli Shechtman, Jeong~Joon Park, and Ira
  Kemelmacher-Shlizerman.
\newblock Stylesdf: High-resolution 3d-consistent image and geometry
  generation.
\newblock \emph{{CVPR}}, 2022.

\bibitem[Park et~al.(2022)Park, Sinha, Hedman, Barron, Bouaziz, Goldman,
  Martin-Brualla, and Seitz]{park2021hypernerf}
Keunhong Park, Utkarsh Sinha, Peter Hedman, Jonathan~T Barron, Sofien Bouaziz,
  Dan~B Goldman, Ricardo Martin-Brualla, and Steven~M Seitz.
\newblock Hypernerf: A higher-dimensional representation for topologically
  varying neural radiance fields.
\newblock \emph{SIGGRAPH}, 2022.

\bibitem[Paysan et~al.(2009)Paysan, Knothe, Amberg, Romdhani, and
  Vetter]{bfm09}
Pascal Paysan, Reinhard Knothe, Brian Amberg, Sami Romdhani, and Thomas Vetter.
\newblock A 3d face model for pose and illumination invariant face recognition.
\newblock In \emph{2009 sixth IEEE international conference on advanced video
  and signal based surveillance}, pp.\  296--301. Ieee, 2009.

\bibitem[Richardson et~al.(2021)Richardson, Alaluf, Patashnik, Nitzan, Azar,
  Shapiro, and Cohen-Or]{richardson2021encoding}
Elad Richardson, Yuval Alaluf, Or~Patashnik, Yotam Nitzan, Yaniv Azar, Stav
  Shapiro, and Daniel Cohen-Or.
\newblock Encoding in style: a stylegan encoder for image-to-image translation.
\newblock In \emph{{CVPR}}, 2021.

\bibitem[Saito et~al.(2017)Saito, Matsumoto, and Saito]{saito2017temporal}
Masaki Saito, Eiichi Matsumoto, and Shunta Saito.
\newblock Temporal generative adversarial nets with singular value clipping.
\newblock In \emph{{ICCV}}, 2017.

\bibitem[Schwarz et~al.(2020)Schwarz, Liao, Niemeyer, and
  Geiger]{schwarz2020graf}
Katja Schwarz, Yiyi Liao, Michael Niemeyer, and Andreas Geiger.
\newblock Graf: Generative radiance fields for 3d-aware image synthesis.
\newblock \emph{{NeurIPS}}, 2020.

\bibitem[Schwarz et~al.(2022)Schwarz, Sauer, Niemeyer, Liao, and
  Geiger]{schwarz2022voxgraf}
Katja Schwarz, Axel Sauer, Michael Niemeyer, Yiyi Liao, and Andreas Geiger.
\newblock Voxgraf: Fast 3d-aware image synthesis with sparse voxel grids.
\newblock \emph{{NeurIPS}}, 2022.

\bibitem[Shen et~al.(2020)Shen, Gu, Tang, and Zhou]{shen2020interpreting}
Yujun Shen, Jinjin Gu, Xiaoou Tang, and Bolei Zhou.
\newblock Interpreting the latent space of gans for semantic face editing.
\newblock In \emph{{CVPR}}, 2020.

\bibitem[Shi et~al.(2021)Shi, Aggarwal, and Jain]{shi2021lifting}
Yichun Shi, Divyansh Aggarwal, and Anil~K Jain.
\newblock Lifting 2d stylegan for 3d-aware face generation.
\newblock In \emph{{CVPR}}, 2021.

\bibitem[Siarohin et~al.(2019)Siarohin, Lathuili{\`e}re, Tulyakov, Ricci, and
  Sebe]{siarohin2019first}
Aliaksandr Siarohin, St{\'e}phane Lathuili{\`e}re, Sergey Tulyakov, Elisa
  Ricci, and Nicu Sebe.
\newblock First order motion model for image animation.
\newblock \emph{{NeurIPS}}, 2019.

\bibitem[Sitzmann et~al.(2020)Sitzmann, Martel, Bergman, Lindell, and
  Wetzstein]{sitzmann2020implicit}
Vincent Sitzmann, Julien Martel, Alexander Bergman, David Lindell, and Gordon
  Wetzstein.
\newblock Implicit neural representations with periodic activation functions.
\newblock \emph{{NeurIPS}}, 2020.

\bibitem[Skorokhodov et~al.(2021)Skorokhodov, Ignatyev, and
  Elhoseiny]{skorokhodov2021adversarial}
Ivan Skorokhodov, Savva Ignatyev, and Mohamed Elhoseiny.
\newblock Adversarial generation of continuous images.
\newblock In \emph{{CVPR}}, 2021.

\bibitem[Skorokhodov et~al.(2022)Skorokhodov, Tulyakov, and
  Elhoseiny]{skorokhodov2021stylegan}
Ivan Skorokhodov, Sergey Tulyakov, and Mohamed Elhoseiny.
\newblock Stylegan-v: A continuous video generator with the price, image
  quality and perks of stylegan2.
\newblock \emph{{CVPR}}, 2022.

\bibitem[Tewari et~al.(2019)Tewari, Bernard, Garrido, Bharaj, Elgharib, Seidel,
  P{\'e}rez, Zollhofer, and Theobalt]{tewari2019fml}
Ayush Tewari, Florian Bernard, Pablo Garrido, Gaurav Bharaj, Mohamed Elgharib,
  Hans-Peter Seidel, Patrick P{\'e}rez, Michael Zollhofer, and Christian
  Theobalt.
\newblock Fml: Face model learning from videos.
\newblock In \emph{{CVPR}}, 2019.

\bibitem[Tian et~al.(2021)Tian, Ren, Chai, Olszewski, Peng, Metaxas, and
  Tulyakov]{tian2021good}
Yu~Tian, Jian Ren, Menglei Chai, Kyle Olszewski, Xi~Peng, Dimitris~N Metaxas,
  and Sergey Tulyakov.
\newblock A good image generator is what you need for high-resolution video
  synthesis.
\newblock In \emph{{ICLR}}, 2021.

\bibitem[Tov et~al.(2021)Tov, Alaluf, Nitzan, Patashnik, and
  Cohen-Or]{tov2021designing}
Omer Tov, Yuval Alaluf, Yotam Nitzan, Or~Patashnik, and Daniel Cohen-Or.
\newblock Designing an encoder for stylegan image manipulation.
\newblock \emph{{ACM Trans. on Graphics}}, 2021.

\bibitem[Tulyakov et~al.(2018)Tulyakov, Liu, Yang, and
  Kautz]{tulyakov2018mocogan}
Sergey Tulyakov, Ming-Yu Liu, Xiaodong Yang, and Jan Kautz.
\newblock Mocogan: Decomposing motion and content for video generation.
\newblock In \emph{{CVPR}}, 2018.

\bibitem[Unterthiner et~al.(2018)Unterthiner, Van~Steenkiste, Kurach, Marinier,
  Michalski, and Gelly]{unterthiner2019fvd}
Thomas Unterthiner, Sjoerd Van~Steenkiste, Karol Kurach, Raphael Marinier,
  Marcin Michalski, and Sylvain Gelly.
\newblock Towards accurate generative models of video: A new metric \&
  challenges.
\newblock \emph{arXiv preprint arXiv:1812.01717}, 2018.

\bibitem[Vondrick et~al.(2016)Vondrick, Pirsiavash, and
  Torralba]{vondrick2016generating}
Carl Vondrick, Hamed Pirsiavash, and Antonio Torralba.
\newblock Generating videos with scene dynamics.
\newblock \emph{{NeurIPS}}, 2016.

\bibitem[Wang et~al.(2021{\natexlab{a}})Wang, Mallya, and
  Liu]{wang2021facevid2vid}
Ting-Chun Wang, Arun Mallya, and Ming-Yu Liu.
\newblock One-shot free-view neural talking-head synthesis for video
  conferencing.
\newblock In \emph{{CVPR}}, 2021{\natexlab{a}}.

\bibitem[Wang et~al.(2021{\natexlab{b}})Wang, Bagautdinov, Lombardi, Simon,
  Saragih, Hodgins, and Zollhofer]{wang2021learning}
Ziyan Wang, Timur Bagautdinov, Stephen Lombardi, Tomas Simon, Jason Saragih,
  Jessica Hodgins, and Michael Zollhofer.
\newblock Learning compositional radiance fields of dynamic human heads.
\newblock In \emph{{CVPR}}, 2021{\natexlab{b}}.

\bibitem[Yu et~al.(2022)Yu, Tack, Mo, Kim, Kim, Ha, and Shin]{yu2022generating}
Sihyun Yu, Jihoon Tack, Sangwoo Mo, Hyunsu Kim, Junho Kim, Jung-Woo Ha, and
  Jinwoo Shin.
\newblock Generating videos with dynamics-aware implicit generative adversarial
  networks.
\newblock In \emph{{ICLR}}, 2022.

\bibitem[Yuan et~al.(2022)Yuan, Sun, Lai, Ma, Jia, and Gao]{yuan2022nerf}
Yu-Jie Yuan, Yang-Tian Sun, Yu-Kun Lai, Yuewen Ma, Rongfei Jia, and Lin Gao.
\newblock Nerf-editing: geometry editing of neural radiance fields.
\newblock In \emph{{CVPR}}, 2022.

\bibitem[Zhang et~al.(2022{\natexlab{a}})Zhang, Jiang, Yang, Xu, Shi, Song, Xu,
  Wang, and Feng]{zhang2022avatargen}
Jianfeng Zhang, Zihang Jiang, Dingdong Yang, Hongyi Xu, Yichun Shi, Guoxian
  Song, Zhongcong Xu, Xinchao Wang, and Jiashi Feng.
\newblock Avatargen: a 3d generative model for animatable human avatars.
\newblock \emph{arXiv preprint arXiv:2211.14589}, 2022{\natexlab{a}}.

\bibitem[Zhang et~al.(2022{\natexlab{b}})Zhang, Zheng, Gao, Zhang, Pan, and
  Yang]{zhang2022multi}
Xuanmeng Zhang, Zhedong Zheng, Daiheng Gao, Bang Zhang, Pan Pan, and Yi~Yang.
\newblock Multi-view consistent generative adversarial networks for 3d-aware
  image synthesis.
\newblock \emph{{CVPR}}, 2022{\natexlab{b}}.

\bibitem[Zhao et~al.(2022)Zhao, Ma, G{\"u}era, Ren, Schwing, and
  Colburn]{zhao2022generative}
Xiaoming Zhao, Fangchang Ma, David G{\"u}era, Zhile Ren, Alexander~G Schwing,
  and Alex Colburn.
\newblock Generative multiplane images: Making a 2d gan 3d-aware.
\newblock In \emph{{ECCV}}, pp.\  18--35. Springer, 2022.

\bibitem[Zhu et~al.(2022)Zhu, Wu, Zhu, Jiang, Tang, Zhang, Liu, and
  Loy]{zhu2022celebvhq}
Hao Zhu, Wayne Wu, Wentao Zhu, Liming Jiang, Siwei Tang, Li~Zhang, Ziwei Liu,
  and Chen~Change Loy.
\newblock {CelebV-HQ}: A large-scale video facial attributes dataset.
\newblock In \emph{{ECCV}}, 2022.

\bibitem[Zhuang et~al.(2022)Zhuang, Ma, Koyejo, and
  Schwing]{zhuang2022controllable}
Peiye Zhuang, Liqian Ma, Oluwasanmi Koyejo, and Alexander~G Schwing.
\newblock Controllable radiance fields for dynamic face synthesis.
\newblock \emph{{3DV}}, 2022.

\end{thebibliography}
